\documentclass[journal]{IEEEtran}
\usepackage[comma,numbers,square,sort&compress]{natbib}
\usepackage{subfigure}
\usepackage{color}
\usepackage{threeparttable}
\usepackage{colortbl}
\usepackage{graphicx}
\usepackage{float}
\usepackage{amsmath,latexsym,amssymb,amsthm,array,amsfonts,algorithm,algpseudocode,booktabs,graphicx,subfigure,multirow,cuted,stfloats}
\usepackage[sort&compress]{natbib}
\usepackage{url}
\usepackage{multirow}
\usepackage{lscape}
\usepackage{nomencl}
\makenomenclature
\usepackage{makecell}
\usepackage{float}
\usepackage{enumerate}
\usepackage{makecell}
\usepackage{bm}
\usepackage{array}
\definecolor{mygray}{gray}{0.8}
\theoremstyle{definition}

\algrenewcommand{\algorithmiccomment}[1]{\hskip1em$/*$ #1 $*/$}

\newlength\savewidth

\newcolumntype{I}{!{\vrule width 1.2pt}}
\newlength\savedwidth

\begin{document}
	\hyphenpenalty=5000
	\tolerance=1200
	\title{Diffusion \textcolor{black}{Model-Based Multiobjective} Optimization for Gasoline Blending Scheduling}
	
	\author{\textcolor{black}{Wenxuan Fang, Wei Du, Renchu He, and Yang Tang \\
			East China University of Science and Technology, CHINA \\
			Yaochu Jin, Westlake University, CHINA \\
			Gary G. Yen, Oklahoma State University, USA }
		% <-this % stops a space
		% \thanks{This work was supported by National Natural Science Foundation of China (Basic Science Center Program: 61988101), National Natural Science Foundation of China (62173144), National Natural Science Foundation of Shanghai (21ZR1416100), Shanghai Rising-Star Program (22QA1402400), Fundamental Research Funds for the Central Universities, Shanghai AI Lab, and in part by the Sino-German Center for Research Promotion under Grant M-0066. Y. Jin is funded by an Alexander von Humboldt Professorship for AI. (Corresponding author: Wei Du.)}
		
		\thanks{\textcolor{black}{Corresponding author: Wei Du (e-mail: duwei0203@gmail.com)}}
		
		%\thanks{W. Fang, W. Du, R. He and Y. Tang are with the Key Laboratory of Smart Manufacturing in Energy Chemical Process, Ministry of Education, East China University of Science and Technology, Shanghai 200237, China (e-mail: y30210943@mail.ecust.edu.cn; duwei0203@gmail.com; renchuhe@ecust.edu.cn; yangtang@ecust.edu.cn).}
		
		%\thanks{Y. Jin is with the School of Engineering, Westlake University, Hangzhou 310030, China (e-mail: jinyaochu@westlake.edu.cn)}
		
		%\thanks{G. Yen is with the School of Electrical and Computer Engineering, Oklahoma State University, Stillwater, OK 74078, USA. (e-mail: gyen@okstate.edu).}
	}

	\markboth{}%
	{Shell \MakeLowercase{\textit{et al.}}: Bare Demo of IEEEtran.cls
		for Journals}

	\maketitle

	\begin{abstract}
		Gasoline blending scheduling \textcolor{black}{uses} resource allocation and operation sequencing to meet a refinery’s production requirements. The presence of nonlinearity, integer constraints, and a large number of decision variables adds complexity to this problem, posing challenges for traditional and evolutionary algorithms. This paper introduces a novel \textcolor{black}{multiobjective} optimization approach driven by \textcolor{black}{a} diffusion model (named DMO), \textcolor{black}{which is designed specifically} for gasoline blending scheduling. To address integer constraints and generate feasible schedules, the diffusion model creates multiple intermediate distributions between Gaussian noise and the feasible domain. Through iterative processes, \textcolor{black}{the} solutions transition from Gaussian noise to feasible schedules while optimizing \textcolor{black}{the} objectives using the gradient descent method. DMO achieves simultaneous objective optimization and constraint adherence. Comparative tests are conducted to evaluate DMO’s performance across various scales. \textcolor{black}{The experimental} results demonstrate that DMO surpasses state-of-the-art \textcolor{black}{multiobjective} evolutionary algorithms in terms of efficiency when solving gasoline blending scheduling problems. 
		%\boldmath
	\end{abstract}
	

	% \fi
	\IEEEpeerreviewmaketitle

	\section{Introduction}
	Gasoline accounts for 40\% of \textcolor{black}{the} overall crude oil production and \textcolor{black}{more than} 70\% of the crude oil product margin. Furthermore, gasoline plays a key role in transportation and industrial production. 
	In recent years, the refinery industry has \textcolor{black}{been facing} daunting challenges such as fluctuating product demands, volatile crude prices, and strict environmental regulations. 
	To thrive in this low-margin business, refineries are increasingly focusing on production optimization rather than investing in new production equipment \cite{WEO2022}. 
	Consequently, optimizing in the gasoline production process will be far more crucial for future prosperity than expanding the production scale. 
	
\begin{figure}[!tbp]
		\centering
		\includegraphics[width=0.49\textwidth]{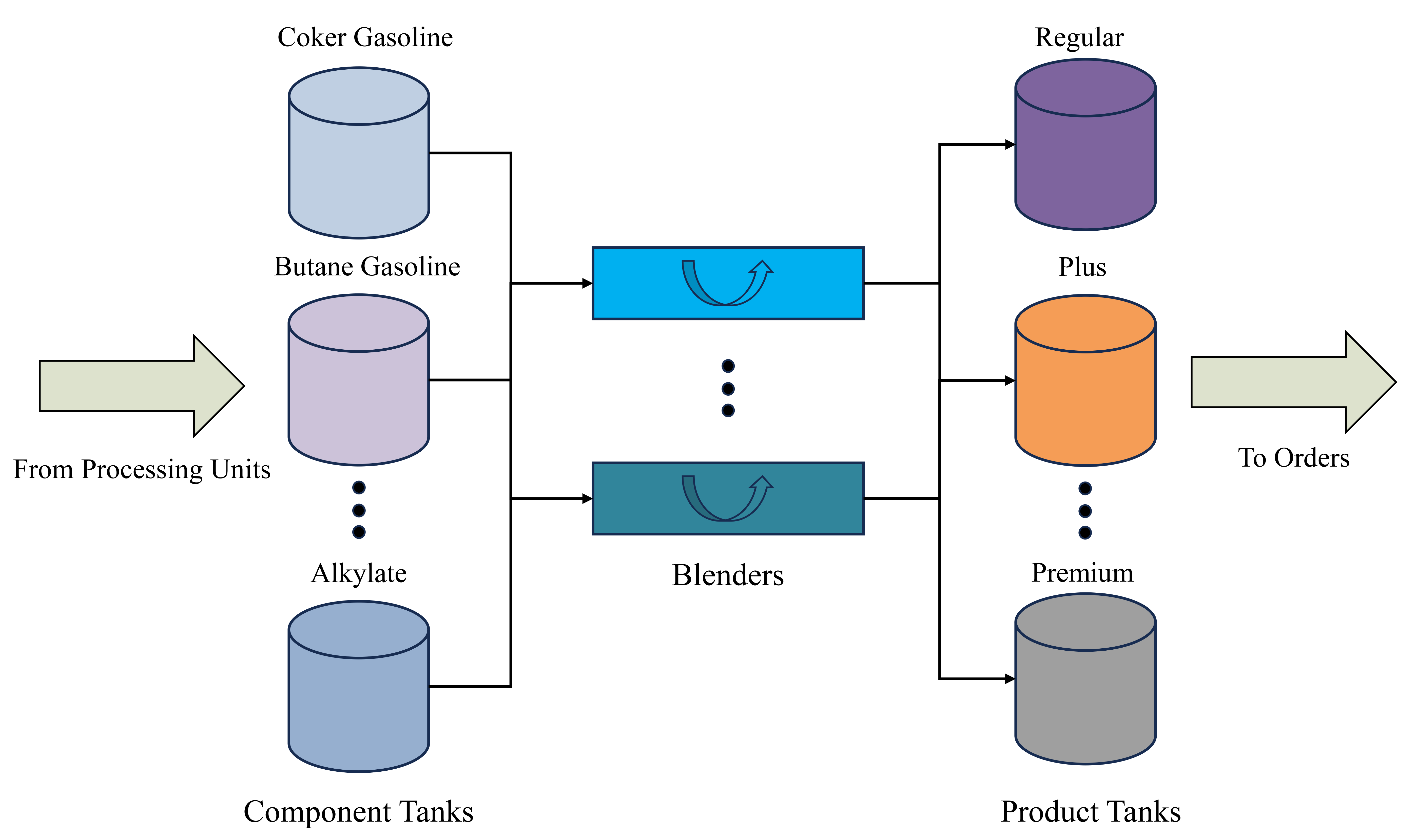}\\
		\caption{A schematic diagram of \textcolor{black}{a} gasoline blending unit. The components from different tanks are blended in varying proportions using the blenders to produce different grades of gasoline. These grades of gasoline are then transferred to product tanks \textcolor{black}{to fulfill} customer orders. 
		}
		\label{fig_Sche}
	\end{figure}

	\textcolor{black}{Gasoline blending is the final step in gasoline production and involves the mixing of components in different proportions to produce various grades of gasoline. Fig. \ref{fig_Sche} shows a schematic diagram of a gasoline blending unit.} Gasoline blending scheduling \textcolor{black}{uses} resource allocation and operation sequencing to meet the refinery's production requirements within specific time frames, typically ranging from a week to a month. An effective schedule should not only meet basic production requirements but also enhance productivity by minimizing unnecessary operations, reducing inventory costs, and minimizing raw material waste. 
	The gasoline blending scheduling problem \textcolor{black}{incorporates} the combinatorial optimization nature of the scheduling problem with the need for accuracy in continuous production. 
	Based on the aforementioned analysis, the gasoline blending scheduling problem can be formulated as a \textcolor{black}{multiobjective} mixed-integer optimization problem encompassing nonlinearity, integer constraints, and a massive number of decision variables. This formulation poses a significant challenge for existing algorithms. 
	
	To address the \textcolor{black}{multiobjective} gasoline blending scheduling optimization problem, two primary methods \textcolor{black}{are} employed: traditional algorithms and evolutionary approaches. \textcolor{black}{For traditional algorithms, numerous attempts have been made to utilize relaxation, branch-and-bound, and cutting-plane techniques to address the gasoline blending scheduling problem.} 
	For instance, Castillo and Castro \cite{LinearScheduling2} employed piecewise McCormick relaxation and a normalized multiparametric disaggregation technique for the global optimization of the gasoline blending scheduling problem. 
	On the other hand, numerous studies have focused on adapting \textcolor{black}{multiobjective} evolutionary algorithms (MOEAs) to \textcolor{black}{address} gasoline blending scheduling problems. 
	For example, Ivanov and Ray \cite{EA1} improved upon MOEAs such as NSGA-II and NAGA \textcolor{black}{by} comparing their performance with \textcolor{black}{that of} existing methods for the gasoline blending scheduling problem. Hou et al. \cite{EA2} proposed an NSGA-III-based optimization algorithm to efficiently find Pareto-optimal solutions. However, their approach is limited to small-scale problems. 
	
	Despite these efforts, the results achieved by traditional algorithms and evolutionary approaches alone have proven unsatisfactory due to several challenges. They struggle with handling small-scale problems and exhibit inconsistent performance when dealing with large-scale scenarios. 
	This issue of scale cannot be resolved by merely switching to MOEAs for large-scale \textcolor{black}{multiobjective} problems, as most MOEAs for such problems group large-scale decision variables based on their relationship to the objectives \cite{LMO, MOEA/DVA}. However, in the gasoline blending scheduling problem, there is no distinction in the roles of the decision variables, as they affect both optimality and diversity. 
	Consequently, current studies aim to address the gasoline blending scheduling problem by imposing additional priorities and assumptions that reduce the complexity \textcolor{black}{of problem definition.} 
	For example, Li and Karimi \cite{LinearScheduling1} revised the schedule adjustment procedure to avoid solving mixed-integer nonlinear programming (MINLP) problems. Panda and Ramteke \cite{SAGA} proposed a structure-adapted genetic algorithm that generates initial schedules using predefined nominal parameters and assesses their robustness in the face of demand uncertainty. Bayu et al. \cite{GGA} developed a graphical genetic algorithm (GGA) model that divides the optimization process into two stages: generating a feasible initial schedule and iteratively refining it. 
	However, the drawback of these algorithms is their lack of flexibility. When faced with different objectives or working environments, they often need to be redesigned from scratch, limiting their applicability to the diverse requirements of industrial production. Hence, there is an urgent need for an easily scalable and stable method to effectively solve the gasoline scheduling problem. 
	
	\textcolor{black}{The above references show that the existing methods often build a mathematical model first and then design an algorithm for optimization. }
	\textcolor{black}{However, when examining a gasoline blending schedule, feasibility can be judged visually from the corresponding image (i.e., the Gantt chart). This is because a schedule must adhere to certain integer constraints, which arise due to the combinatorial nature of scheduling problems.  }
	For instance, \textcolor{black}{simultaneous receiving and delivery of product} is not allowed for a single oil tank, and frequent switching of operations should be minimized. 
	Therefore, we \textcolor{black}{aim} to solve the gasoline blending scheduling problem by generating images that match the schedule characteristics instead of building a mathematical model with complex constraints. 
	This observation has motivated us to exploit the use of diffusion models known for their successful performance in image synthesis. 
	
	The diffusion model is specifically designed to model complex datasets using a highly flexible family of probability distributions. It has shown promising results in various areas, including image restoration and data synthesis \cite{DM}. 
	Following the successful utilization of the diffusion model in image synthesis by DALLE-2 \cite{DALLE-2}, it has quickly surpassed generative adversarial networks (GANs) \cite{BeatGans} and \textcolor{black}{became} the most powerful deep generative model, exhibiting remarkable performance across multiple applications such as image synthesis \cite{LDM}, video generation \cite{DMgVideo}, and audio synthesis \cite{DMgAudio}. In recent years, a multitude of improved \textcolor{black}{variations have} been proposed to enhance speed-up improvements \cite{IDM}, strengthen generative capacity \cite{MoreControl}, and employ dimension reduction techniques \cite{LDM}. Leveraging the flexibility and strength of diffusion models, they have been increasingly employed to \textcolor{black}{address} a wide range of challenging real-world tasks, including molecular graph modeling \cite{MGD1, MGD2} and material design \cite{MD1, MD2}. 
	\textcolor{black}{However, studies on the application of diffusion models to optimization problems are still limited \cite{DIFUSCO}. }
	
	\textcolor{black}{Based on the above discussion, in this paper, we develop a \textcolor{black}{multiobjective} optimization method driven by a diffusion model, referred to as DMO, to solve the gasoline blending scheduling problem directly.} Additionally, the proposed DMO can be easily extended to other types of scheduling problems or combinatorial optimization problems. 
	In the proposed DMO, to generate a solution that meets integer constraints, namely, a schedule \textcolor{black}{that is} considered feasible, the diffusion model creates multiple intermediate distributions between Gaussian noise and the feasible domain. 
	Then, \textcolor{black}{during} the iteration of the diffusion model, \textcolor{black}{the} solution gradually changes from Gaussian noise to a schedule considered feasible, while the objectives are optimized by the gradient descent method. 
	Therefore, DMO can simultaneously optimize the objectives and ensure that \textcolor{black}{the} solutions adhere to the constraints. 
	
	The contributions of this paper can be summarized as follows: 
	\begin{itemize}
		\item [1)] 
		This study represents the first attempt to employ diffusion models in \textcolor{black}{solving} a practical scheduling optimization problem, specifically the gasoline blending scheduling problem. In particular, the proposed DMO successfully addresses the challenges posed by integer constraints and the vast search space associated with the gasoline blending scheduling problem. This achievement is made possible by harnessing the powerful generative capabilities inherent in diffusion models. 
		\item [2)]
		DMO introduces a novel optimization path by generating multiple intermediate distributions between Gaussian noise and the feasible domain. This approach enables DMO to simultaneously optimize conflicting objectives while ensuring adherence to constraints. Consequently, DMO exhibits a robust search capability within the solution space. 
		\item [3)]
		Extensive experiments demonstrate that DMO offers significant improvements in terms of stability and efficiency when applied to the gasoline blending scheduling problem. In comparison to state-of-the-art MOEAs, DMO consistently delivers superior and more stable solutions. 
	\end{itemize}
	
	The remainder of this paper is organized as follows. Section II presents the background of the study. Section III provides the problem formulation of the gasoline blending scheduling problem investigated in this paper. Section IV introduces the proposed DMO in detail. In Section V, extensive simulation studies are carried out to validate the effectiveness of DMO. Finally, the paper is concluded in Section VI, \textcolor{black}{which provides pertinent observations. }
	
	\section{\textcolor{black}{Background and Motivation}}
	\textcolor{black}{In this section, the background of gasoline mixing scheduling and diffusion modeling is presented, as well as the motivation behind DMO. }
	
	\subsection{Gasoline Blending Scheduling}
	
	\textcolor{black}{Gasoline production is a complex process and involves both physical and chemical processes.} 
	It primarily consists of separation, conversion, and treatment \textcolor{black}{steps}. The treatment \textcolor{black}{step} specifically focuses on gasoline blending. \textcolor{black}{To} produce gasoline with predetermined properties, the components are mixed in varying proportions using a blender, based on product specifications and other requirements. This ensures that the gasoline exhibits an even combustion pattern, complies with current environmental regulations, and \textcolor{black}{facilitates combustion initiation under extreme weather conditions}. Once blended, it is crucial to distribute \textcolor{black}{the} gasoline promptly to meet customer demands in terms of both timeliness and quantity. Gasoline blending, \textcolor{black}{being} the final step, directly influences product quality and overall plant efficiency. 
	
\begin{figure}[tp]
		\centering
		\subfigure[]{
			\begin{minipage}[b]{0.45\textwidth}
				\includegraphics[width=1\textwidth]{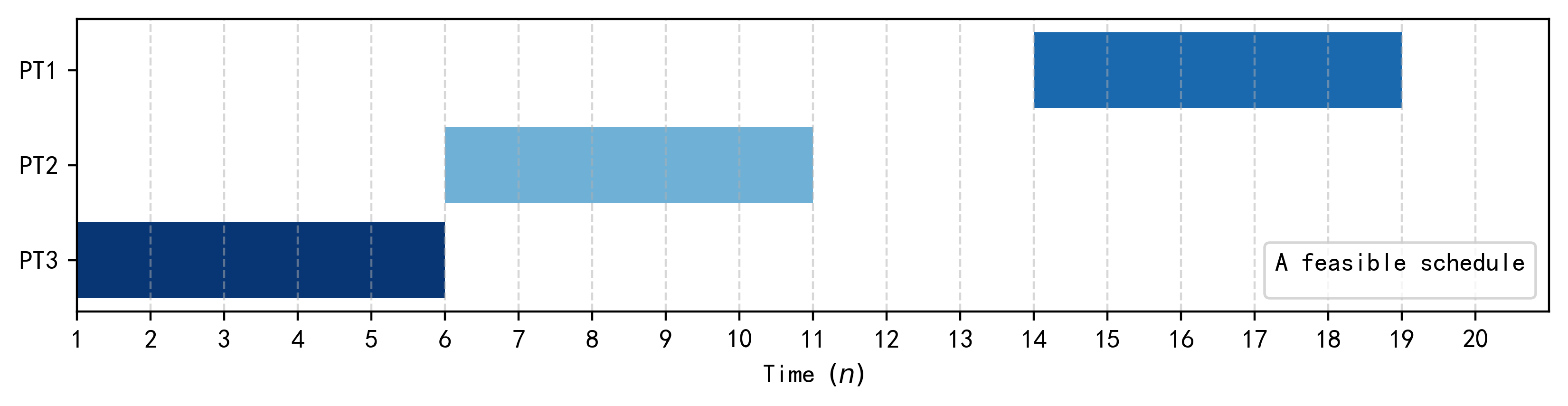}
			\end{minipage}\label{fig_ScheduleA}
		}
		\qquad
		\subfigure[]{
			\begin{minipage}[b]{0.45\textwidth}
				\includegraphics[width=1\textwidth]{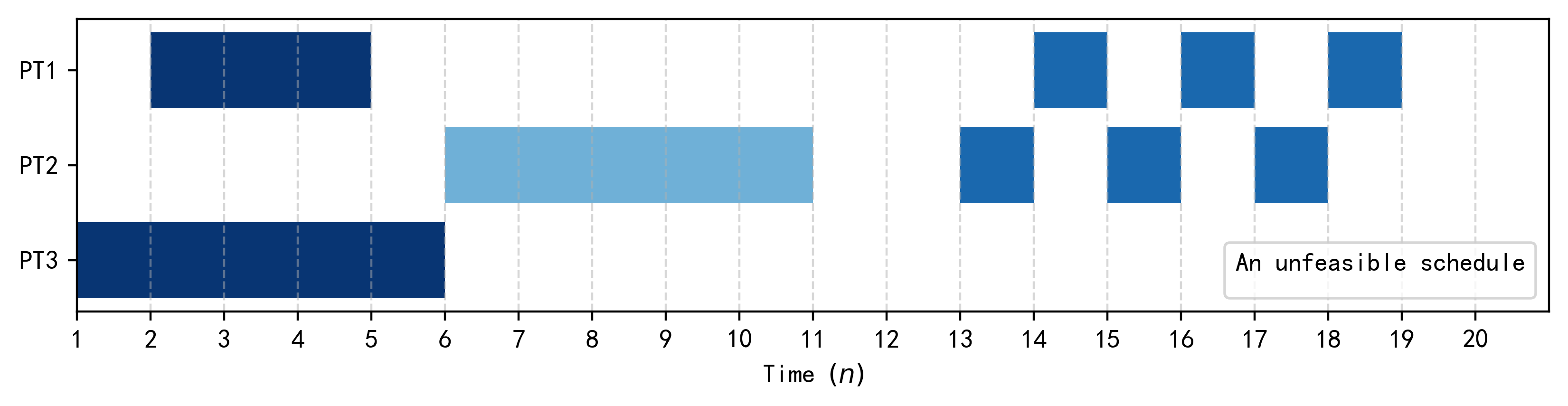}
			\end{minipage}\label{fig_ScheduleB}
		}
		\caption{\textcolor{black}{Examples of the Gantt charts. The colored areas indicates that an oil component tank is transferring oil to product tank $j$ (vertical coordinate) during this specific period (horizontal coordinate).} The shades of blue indicate the flow of oil transportation. (a) A feasible schedule. (b) An unfeasible schedule.} \label{fig_Schedule}
	\end{figure}

	The refining industry is generally considered to \textcolor{black}{use} continuous production, which requires \textcolor{black}{the optimization of} continuous variables. However, due to the uncertainty of production materials and demand, gasoline blending scheduling inevitably \textcolor{black}{involves} combinatorial optimization, similar to ordinary scheduling problems. This introduces various integer constraints that can be observed directly from the schedule \textcolor{black}{in a} Gantt chart. 
	\textcolor{black}{For example, in Fig. \ref{fig_ScheduleA}, a feasible schedule is represented, while Fig. \ref{fig_ScheduleB} depicts an unfeasible schedule.}
	In these figures, a colored area indicates that an oil component tank is transferring oil to product tank $j$ (vertical coordinate) during a specific period (horizontal coordinate). The shades of blue indicate the flow of oil transportation. 
	\textcolor{black}{In Fig. \ref{fig_ScheduleA}, the given component tank has three tasks: delivering oil to product tank 3 during time periods 1-5, to product tank 2 during time periods 6-10, and to product tank 1 during time periods 14-18. These tasks do not conflict with each other and can be easily executed. }
	In Fig. \ref{fig_ScheduleB},  there is a conflict in operation (one component tank delivers oil to two product tanks simultaneously) in the first half, and the operations in the second half are too fragmented to be executed. 
	
	For traditional algorithms, gasoline blending scheduling is formulated as a mixed-integer programming (MIP) and falls into the class of NP-hard problems. 
	Exact methods \textcolor{black}{for solving these problems}, such as branch-and-bound and cutting-plane \textcolor{black}{methods}, are computationally expensive and not suitable for solving practical problems \cite{MIP}. 
	On the other hand, evolutionary algorithms also struggle to efficiently handle integer constraints. 
	For instance, genetic algorithms cannot guarantee that the offspring generated from the crossover operator will satisfy integer constraints, resulting in a significant number of wasted iterations. Similar issues arise with other evolutionary algorithms, as they all rely on random methods to generate children from parents. 
	\textcolor{black}{Therefore, the primary challenge in crude oil reconciliation and scheduling is to generate solutions that meet the specified requirements (schedules that are considered feasible) in the form of Gantt charts. }
	This situation brings to mind diffusion models, \textcolor{black}{which have proven effective in image synthesis. }
	
	\subsection{Diffusion Models}
	
	\textcolor{black}{Image synthesis is the task of generating new images from an existing dataset. Over the last decade, a wide variety of training data generation methods have been proposed.} In particular, state-of-the-art GANs \cite{GANS} \textcolor{black}{can} generate high-fidelity natural images \textcolor{black}{in} diverse categories. GANs are architectures for automatically training a generative model by treating the unsupervised problem as supervised and using both a generator and a discriminator model. \textcolor{black}{The generator attempts to create realistic samples that deceive the discriminator, while the discriminator strives to distinguish real samples from fake samples.} However, dynamically training two models makes GANs difficult to train, often resulting in collapse without carefully selected hyperparameters \textcolor{black}{or} regularization. 
	
\begin{figure}[!tbp]
		\centering
		\includegraphics[width=0.49\textwidth]{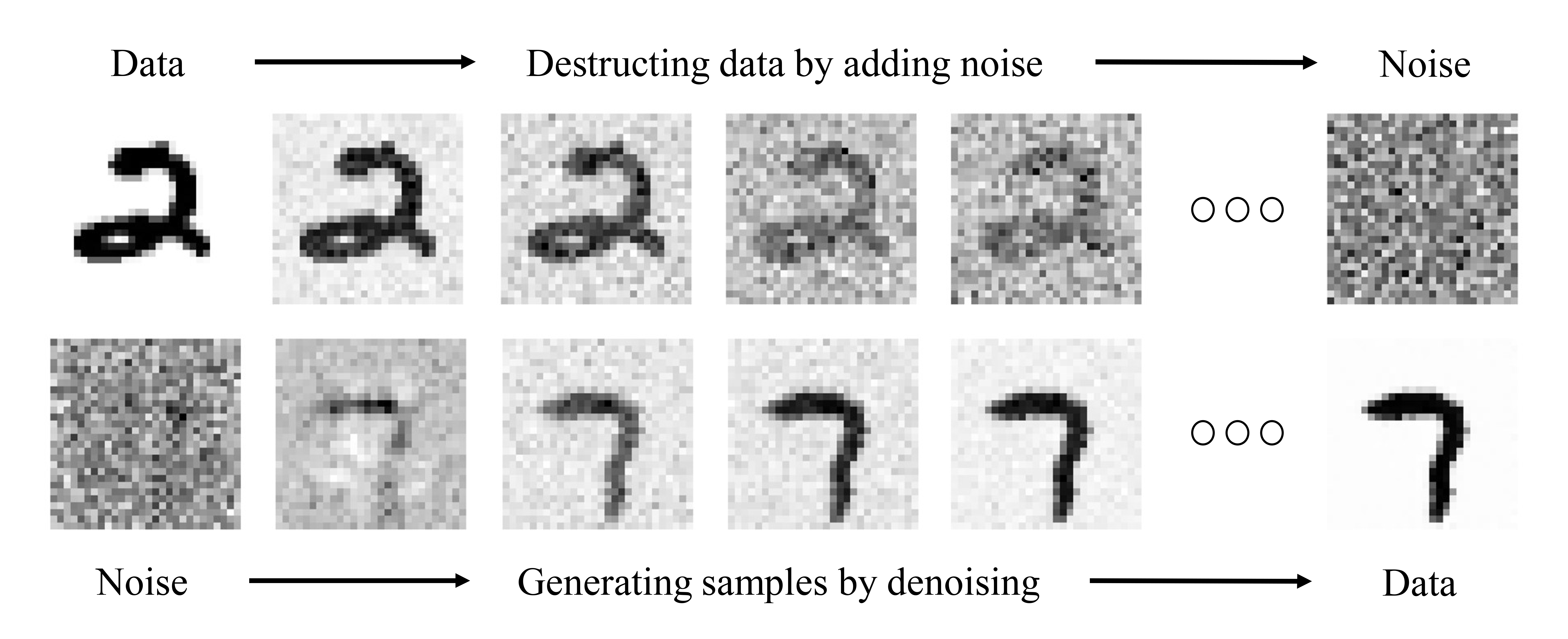}\\
		\caption{Illustration of how diffusion models operate on \textcolor{black}{the} MNIST handwritten digit database. 
		}
		\label{fig_DMdigit}
	\end{figure}

	Diffusion models abandon the idea of finding the original data distribution in one step and instead, create several intermediate distributions between the Gaussian noise and the desired probability distribution, thereby enabling data generation. Although diffusion models are capable of more than image synthesis, as the schedule to be generated in this study can be considered a special kind of picture, we will focus on how diffusion models generate images. Fig. \ref{fig_DMdigit} demonstrates how diffusion models operate on \textcolor{black}{the} MNIST handwritten digit database \cite{MNIST}. The first line adds Gaussian noise to \textcolor{black}{an} image, while the second line employs a denoising model to progressively eliminate the noise and generate \textcolor{black}{an} image. As a result of the incremental image generation process, diffusion models exhibit improved stability. Additionally, the step-by-step approach enhances scalability. \textcolor{black}{By iterating over images using a classifier, images with specific content can be generated, as exemplified by the generation of the digit 7 in Fig. \ref{fig_DMdigit}. }
	
	Although diffusion models were initially designed to generate data distributions \cite{DM}, recent research on diffusion models has focused \textcolor{black}{primarily} on image synthesis, with most improvements tailored to the specific characteristics of images \cite{IDM}. 
	An important enhancement involves transforming images into the latent space before processing them with the diffusion model, addressing the efficiency issues \textcolor{black}{that were} previously criticized \cite{LDM}. Furthermore, to generate richer content and style, gradients of image-text or image matching scores are utilized to guide image synthesis \cite{MoreControl}. 
	However, research into the application of diffusion models for optimization problems is extremely limited, and the potential of diffusion models for optimization remains largely unexplored. 
	
	\subsection{\textcolor{black}{Motivation}}
	
	Generative models have a long history in traditional machine learning. \textcolor{black}{Many research have been attempting to apply these methods to evolutionary optimization \cite{LEA}.} For example, LGSEA \cite{GTMmoea} captures regularity via a classical machine learning method called generative topographic mapping (GTM) \cite{GTM} to guide evolutionary search to generate promising offspring solutions. 
	\textcolor{black}{Additionally, generative models, such as Bayesian networks, are utilized in estimation of distribution algorithms (EDAs), }sampling new solutions using a probabilistic model derived from statistics extracted from the existing solutions to mitigate the adverse impact of genetic operators \cite{BAYES}. 
	In recent years, the advancement of deep learning has significantly enhanced \textcolor{black}{the generative capabilities of these algorithms,} leading to the development of sophisticated deep generative models such as VAEs \cite{VAE}, GANs \cite{GANS}, and Flow \cite{flow}. However, their application to optimization problems remains relatively limited \cite{GMOEA, GMOEA2, GMOEA3, GMOEA4}. Previous studies have \textcolor{black}{focused primarily} on integrating GANs into evolutionary optimization. 
	For example, GANs have been utilized to generate promising offspring solutions for \textcolor{black}{multiobjective} evolutionary algorithms \cite{GMOEA}.
	In another study \cite{GMOEA2}, a GAN-based manifold interpolation framework was developed to generate high-quality solutions on a manifold for large-scale \textcolor{black}{multiobjective} optimization.
	Despite these efforts, these models typically comprise basic architectures with only a few fully connected layers, significantly lagging behind the \textcolor{black}{sophistication} of contemporary deep learning networks. 
	The primary reason for this limitation is that these methods \textcolor{black}{involve training} on specific data distributions, directly generating approximated new data. However, when aiming to create specific features, additional discriminators become necessary. This approach \textcolor{black}{has} two drawbacks when applied to real optimization problems: 
	
	\begin{itemize}
		\item [1)] 
		\textcolor{black}{The aforementioned generative models output \textcolor{black}{their} results directly in a single step, while optimization algorithms typically operate in stepwise, iterative processes. The disparity between these approaches complicates their efficient integration. }
		\item [2)]
		\textcolor{black}{The discriminators on which generative models rely \textcolor{black}{to generate} data with specific features must be trained on the target data. However, in real optimization problems, the distribution of optimal solutions is unknown in advance, hindering pretraining of the models. Training these models during application of the optimization algorithm would render the algorithm inefficient, as they typically require at least a few hours of training time. However, simplifying the model to reduce training time would significantly degrade the model's generative capacity. }
	\end{itemize}
	
	Diffusion models generate data through a step-by-step iterative approach, \textcolor{black}{which is} perceived as less efficient for image generation problems. However, in optimization problems, this approach can overcome the aforementioned drawbacks. \textcolor{black}{During the iteration of the diffusion model, alternative methods can be implemented to optimize the objective value.} Consequently, diffusion models need to learn \textcolor{black}{only} the distribution of general solutions, allowing for \textcolor{black}{pretraining} of the model. Moreover, considering the gasoline blending scheduling problem's solution as a scheduling Gantt chart possessing image-like properties, a diffusion model can generate this solution. This reasoning motivates our application of the diffusion model to the gasoline blending scheduling problem. 
	
	\section{Problem Formulation}
	\textcolor{black}{In this section, the gasoline blending scheduling model used in this work is introduced. }
	The modeling of gasoline blending scheduling can be divided into discrete- and continuous-time methods. \textcolor{black}{Continuous-time} models are \textcolor{black}{more realistic}, but they also increase the level of complexity in defining the problem, particularly in determining the decision variables \cite{continuous}. 
	The discrete-time model divides time into time periods and requires each change to occur at the beginning and end of the time period. 
	Therefore, in this study, the gasoline blending scheduling problem is modeled using discrete time. 
	
	The model constructed in this paper takes into account the following assumptions. 
	
	\textbf{Assumptions}
	\begin{itemize}
		\item [1)]
		The mixing in a blender is complete, which means that the product properties are uniform and constant. 
		\item [2)]
		All operations start at the beginning of a planning period and terminate at the end of a planning period. 
		\item [3)]
		Tasks must be completed within the scheduling horizon, and the demanded amounts should be fulfilled. 
		\item [4)]
		There are no changes in the properties of the oil in the component tank, and the property data \textcolor{black}{are} accurate without measurement errors. 
		\item [5)]
		The blending of all the properties is linear. 
		\item [6)]
		The target production volume will not exceed the capacity of the product tank. 
	\end{itemize}
	
	\subsection{Sets and variables}
	The following sets are introduced for the gasoline blending scheduling model. 
	\begin{itemize}
		\item $t\in \{t_1,t_2,...,t_n\}$ for $n$ planning periods. As mentioned above, \textcolor{black}{the} gasoline blending scheduling model in this paper is formulated based on a discrete-time representation, where the schedule horizon is divided into several equal time intervals. In this study, the interval is set to two hours. This choice takes into consideration the time-consuming process of opening and closing valves and pumps in the refinery, making two hours a more appropriate interval. 
		\item $N_{ct}$ for the number of component tanks. 
		\item $N_{pt}$ for the number of product tanks. 
		\item $i\in I$ for the index of component tanks. 
		\item $j\in J$ for the index of product tanks. 
		\item $k\in K$ for the property index of gasoline. 
	\end{itemize}
	
	The following \textcolor{black}{are} the decision variables of the gasoline blending scheduling model.
	\begin{itemize}
		\item $ W_{i,j,t} $ for the assignment of gasoline transfer from component tank $i$ to product tank $j$ during period $t$. \textcolor{black}{This is a binary variable.}
		\item $ Q_{i,j,t} $ for the amount of gasoline \textcolor{black}{transferred from component tank $i$ to product tank $j$} during period $t$.
	\end{itemize} 
	
	The following \textcolor{black}{are} the environmental variables of the gasoline blending scheduling model.
	\begin{itemize}
		\item $V'_{i,0}$ for the initial inventory in component tank $i$.
		\item $V'_{min,i}, V'_{max,i}$ for the capacity \textcolor{black}{limits} of component tank $i$.
		\item $Q_{min}, Q_{max}$ for the minimum and maximum allowable flow \textcolor{black}{rates, respectively}, from component tank $i$.
		\item $W'_{i,t}$ for whether component tank $i$ is occupied during period $t$. \textcolor{black}{This is a binary variable.}
		\item $W''_{j,t}$ for whether product tank $j$ is occupied during period $t$. \textcolor{black}{This is a binary variable.}
		\item $V_{j}$ for amount \textcolor{black}{of gasoline} that product tank $j$ needs during the entire optimization time period.
		\item $\Delta C_{i,j,k}$ for the difference of the ratio of property $k$ of gasoline in component tank $i$ and the ratio of property $k$ that  product tank $j$ needs.
	\end{itemize}
	
	The values of these variables do not change during the optimization process. However, the environmental variables differ in each optimization execution cycle and have a significant impact on the difficulty and the results of the problem.
	
	\subsection{Constraints}
	\begin{itemize}
		\item [1)]
		\textbf{Operation constraints}: At any given time, a component tank can only supply oil to one product tank. The tanks may also be occupied by other operations. When a tank is occupied, the component tank cannot send oil, and the product tank cannot receive oil.

		\begin{equation}
			\begin{aligned}
				\forall i,t\ \ W'_{i,t} + \sum_jW_{i,j,t}\le1 \\
				\forall j,t\ \ W''_{j,t} \cdot \sum_iW_{i,j,t}\le1
			\end{aligned}
			\label{eq2.1}
		\end{equation}
		
		\item [2)]
		\textbf{Capacity constraints}: The oil level in the tanks must be maintained within the minimum and maximum limits for safety \textcolor{black}{purposes}. 
		
		\begin{equation}
			\begin{aligned}
				\forall i,t\  V'_{min,i}\le V_{i,t}\le V'_{max,i}
			\end{aligned}
			\label{eq2.2}
		\end{equation}
		
		Since it is assumed that the target production volume will not exceed the capacity of the product tank, only the component tanks are taken into consideration. 
		
		\item [3)]
		\textbf{Flow rate constraints}: The flow rates in the pipelines must be controlled within the maximum and minimum limits. 
		\begin{equation}
			\begin{aligned}
				\forall i,j,t\  Q_{min}\le Q_{i,j,t}\le Q_{max}
			\end{aligned}
			\label{eq2.3}
		\end{equation}
		
		\item [4)]
		\textbf{Operation constraints}: The number of switching operations is limited to a specified \textcolor{black}{value} $N_{max}$. These constraints help minimize unnecessary operations and facilitate the execution of the schedule. The value of N is related to the scale of the problem, and is set to $0.5 \times N_{ct} \times n$.
		
		\begin{equation}
			\begin{aligned}
				\sum_i\sum_j\sum_{t=1}^{n-1}(W_{i,j,t}\oplus W_{i,j,t+1}) \le N_{max}
			\end{aligned}
			\label{eq2.4}
		\end{equation}
		
	\end{itemize}
	
	For the above constraints, the flow rate constraints can be treated as intervals for the decision variables. 
	
	\subsection{Objectives}
	
	Considering the actual requirements, two objectives are set to ensure the quality and feasibility of the final set. 
	
	\begin{itemize}
		\item [1)]
		\textbf{The first objective: minimizing the blending error in gasoline blending, }
		\begin{equation}
			\begin{aligned}
				\min E_{blend} = \sum_{t,j,k} (\sum_i W_{i,j,t}Q_{i,j,t}\Delta C_{i,j,k})^2\\ 
			\end{aligned}
			\label{eq2.5}
		\end{equation}

		The objective function \textcolor{black}{in} Eq. (\ref{eq2.5}) is used to calculate the mean squared deviation of the properties of gasoline, such as the octane number (ON), Reid vapor pressure (RVP), and sulfur content. This objective ensures that the quality of the product meets the required specifications. 
		
		\item [2)]
		\textbf{The second objective: minimizing the yield error of gasoline blending, }
		\begin{equation}
			\begin{aligned}
				\min E_{yield} =&\sum_j(\sum_t\sum_i W_{i,j,t}Q_{i,j,t} - V_j)^2 
			\end{aligned}
			\label{eq2.6}
		\end{equation}
		
		The objective function \textcolor{black}{in} Eq. (\ref{eq2.6}) is used to calculate the mean squared deviation between the actual and the \textcolor{black}{demanded} yield. The schedule must \textcolor{black}{enable production of} a specified quantity of one or more types of gasoline within the scheduling horizon to accomplish this objective. This objective ensures that the production volume of gasoline aligns with the planned targets. 
		
	\end{itemize}
	
	\section{The Proposed DMO} 
	
	\subsection{The Overall Framework of DMO}
	
	The framework of DMO is outlined in Algorithm \ref{Algo}. 
	The input to DMO \textcolor{black}{is} a denoising model $(\mu_\theta(x_t), \Sigma_\theta(x_t))$ trained by images of schedules from historical data, along with \textcolor{black}{the} parameter $s$, which controls the gradient scale and weights of objectives $\textbf{w}$. 
	DMO starts with an initial population $X_T$ sampled from a standard normal distribution (Line 1). Then, $X_t$ iterates through the model $(\mu_\theta(x_t), \Sigma_\theta(x_t))$. 
	During this process, $X_t$ is repeatedly sampled from a distribution determined by a diffusion model and the gradient of \textcolor{black}{the} objectives (Lines 2-6). 
	By imposing the gradient $g$ that drives the objectives down toward the expectation $\mu$, DMO can optimize \textcolor{black}{the} objectives while generating scheduling Gantt charts. 
	The parameter $s$ controls the overall scale of the gradient, while the weights \textbf{w} control the distribution of the solution in the target space to obtain the complete Pareto front. 
	\textcolor{black}{Finally, \textcolor{black}{nondominated} solutions are obtained from $X_1$ after standardization as the output of the algorithm (lines 7-8).
		The overall flowchart of DMO, illustrated in Fig. 4, reveals that the tasks of optimizing the objectives and aligning the solutions with the constraints are intertwined. This integration represents a significant departure from existing optimization methods. }
	
	{\linespread{1.0}
		\begin{algorithm}[!ht]
			\caption{The overall framework of DMO}
			\label{alg_overall_framework}
			\hspace*{0.05in} {\bf Input:} $(\mu_\theta(x_t), \Sigma_\theta(x_t))$ -- a well-trained denoising model\\  
			\hspace*{0.46in} $s$ -- gradient scale\\
			\hspace*{0.46in} $\textbf{w}$ -- weights of objectives
			\begin{algorithmic}[1]
				\Ensure{$P$, the Pareto front}
				\State $X_T \leftarrow$ sample from $\mathcal{N}(\textbf{0}, \textbf{I})$
				\State for all $t$ from $T$ to $2$ do
				\State \ \   $\mu,\Sigma \leftarrow \mu_\theta(X_t), \Sigma_\theta(X_t)$
				\State \ \   $g \leftarrow \nabla_{X_t}[\textbf{w}\cdot (E_{blend}, E_{yield})+ E_{const}]$ 
				\State \ \   $X_{t-1} \leftarrow$ sample from $\mathcal{N}(\mu_\theta(X_t) + s\cdot g, \Sigma_\theta(X_t))$
				\State end for
				\State Standardized $X_1$ and obtain \textcolor{black}{nondominated} solutions $P$
				\State \Return $P$
			\end{algorithmic}
			\label{Algo}
		\end{algorithm}
	}
	
	\begin{figure}[b]
		\centering
		\includegraphics[width=0.49\textwidth]{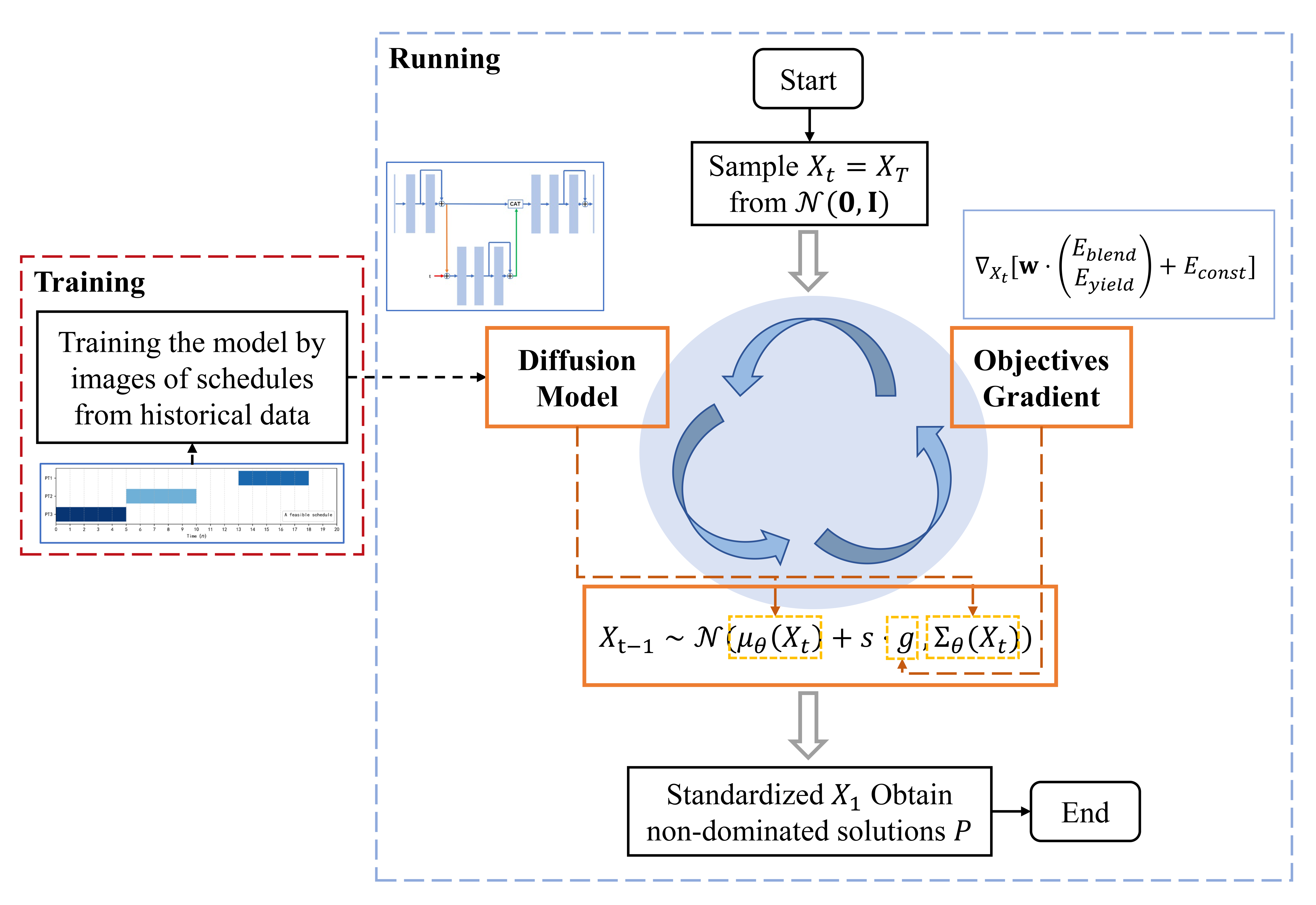}\\
		\caption{The overall flowchart of DMO. 
		}
		\label{fig_FC}
	\end{figure}
	
	\textcolor{black}{In the following subsection, the specific components of DMO are introduced. }
	
	\subsection{The Diffusion Model in DMO}
	
	\subsubsection{Model Analysis}
	
	The core component of DMO is the diffusion model. The mechanism of diffusion models is typically elucidated from the standpoint of variational inference \cite{DM, VDM}. Given that the main objective of DMO is to apply the diffusion model to a practical optimization problem, this introduction will primarily emphasize \textcolor{black}{the aspects of implementation} rather than the underlying principles. 
	
	The diffusion model can be divided into forward and reverse processes. 
	
	\begin{itemize}
		\item 
		\textbf{Forward process}: Given a data distribution $x_0\sim q(x)$ (images of the schedule), Gaussian noise is incrementally added to the sample in $T$ steps, producing latent \textcolor{black}{values} $x_1$ through $x_T$. The forward process in continuous diffusion is defined as follows:
		\begin{equation}
			\begin{aligned}
				x_t = \sqrt{1 - \beta_t} x_{t-1} + \sqrt{\beta_t}\epsilon_{t-1}
			\end{aligned}
			\label{eq3.1}
		\end{equation}
		Here, $\beta_t$ represents the corruption ratio and $\epsilon_{t-1} \sim \mathcal N(\textbf{0}, \textbf{I})$. The general formula derived from this is denoted as follows: 
		\begin{equation}
			\begin{aligned}
				x_t = \sqrt{\overline{\alpha}_t} x_0 + \sqrt{1-\overline{\alpha}_t}\epsilon
			\end{aligned}
			\label{eq3.2}
		\end{equation}
		where $\alpha_T = 1-\beta_t$ and $\overline{\alpha}_t = \prod_{\tau=1}^t \alpha_\tau $. 
		Using Eq. (\ref{eq3.2}), $x_t$ can be easily calculated. 
		In this study, the cosine schedule for $\overline{\alpha}_t$ \cite{IDM} \textcolor{black}{is} utilized: 
		\begin{equation}
			\begin{aligned}
				\overline{\alpha}_t= \frac{f(t)} {f(0)},\  f(t)=\cos {(\frac {t/T+s}{1+s}\cdot \frac \pi 2 )^2}
			\end{aligned}
			\label{eq3.3}
		\end{equation}
		where $s$ is often set to 0.008 to prevent $\beta_t$ from becoming too small near $t=0$.

		\item 
		\textbf{Reverse process}: \textcolor{black}{If there is a recurrence relation formula for deriving $x_{t-1}$ from $x_t$, we can sample $x_T \sim \mathcal N(\textbf{0}, \textbf{I})$ and run this formula repeatedly until we obtain a sample $x_0\sim q(x)$. }
		This formula can be obtained using Bayes’ theorem and Eq. (\ref{eq3.2}): 
		\begin{equation}
			\begin{aligned}
				x_{t-1} &\sim \mathcal{N} (\tilde{\bf\mu}(x_t,\epsilon), \tilde{\beta}_t\bf{I}) \\
				\text{where }\ \tilde{\bf\mu}(x_t,\epsilon) &= \frac {1} {\sqrt{\alpha_t}}(x_t-\frac{1-\alpha_t}{1-\overline\alpha_t}\epsilon)\\ \tilde{\beta}_t&=\frac{1-\overline\alpha_{t-1}}{1-\overline\alpha_t} \beta_t
			\end{aligned}
			\label{eq3.4}
		\end{equation}
		Since the Gaussian noise $\epsilon$ is the only unknown term in Eq. (\ref{eq3.4}), we train the denoising neural network to predict $\epsilon=f_\theta(x_t,t)$.  The loss function is defined as follows: 
		\begin{equation}
			\begin{aligned}
				Loss = E_{t,x_0,\epsilon}(||\epsilon-f_\theta(x_t,t)||^2)
			\end{aligned}
			\label{eq3.5}
		\end{equation}
		
	\end{itemize}
	
	\subsubsection{Model Training}
	
	The purpose of using diffusion models for the gasoline blending scheduling problem is to generate a schedule image that satisfies specified requirements. Hence, the diffusion model is trained \textcolor{black}{on} schedule images from the historical data of a refinery. For each batch, different schedule images $x_0$ are selected from the historical data. Using Eq. (\ref{eq3.2}) and known Gaussian noise $\epsilon$, $x_t$ is obtained as input for the model $f_\theta(x_t,t)$. Then, an optimizer is used to train the model based on the loss function in Eq. (\ref{eq3.5}). Finally, the above steps are repeated until convergence. The training process of the diffusion model is \textcolor{black}{described} in Algorithm \ref{alg_training}. 
	
	{\linespread{1.0}
		\begin{algorithm}[!ht]
			\caption{Model Training}
			\label{alg_training}
			\hspace*{0.05in} {\bf Input:} $f_\theta(x_t)$ -- an untrained denoising model\\
			\hspace*{0.46in} $q(x_0)$ -- the training data
			\begin{algorithmic}[1]
				\Ensure{$(\mu_\theta(x_t), \Sigma_\theta(x_t))$ -- a well-trained denoising model}
				\State \textbf{repeat}
				\State \ \ \ \   $x_0 \leftarrow$ sample from $q(x_0)$
				\State \ \ \ \   $t \leftarrow$ sample from Uniform$({1,...,T})$
				\State \ \ \ \   $\epsilon \leftarrow$ sample from $\mathcal{N}(\textbf{0}, \textbf{I})$
				\State \ \ \ \   $x_t \leftarrow \sqrt{\overline{\alpha}_t} x_0 + \sqrt{1-\overline{\alpha}_t}\epsilon$
				\State \ \ \ \   Take gradient descent step on
				\State \ \ \ \ \ \ \ \ \ \  $\nabla_\theta||\epsilon-f_\theta(x_t,t)||^2$
				\State \textbf{until} converged
				\State $\mu_\theta(x_t)\leftarrow \frac {1} {\sqrt{\alpha_t}}(x_t-\frac{1-\alpha_t}{1-\overline\alpha_t}f_\theta(x_t,t)) $
				\State $\Sigma_\theta(x_t) \leftarrow \frac{1-\overline\alpha_{t-1}}{1-\overline\alpha_t} \beta_t $ 
				\State \Return $(\mu_\theta(x_t), \Sigma_\theta(x_t))$
			\end{algorithmic}
			\label{alg_training_}
		\end{algorithm}
	}
	
	\subsubsection{Network Architecture}
	
\begin{figure}[b]
		\centering
		\includegraphics[width=0.49\textwidth]{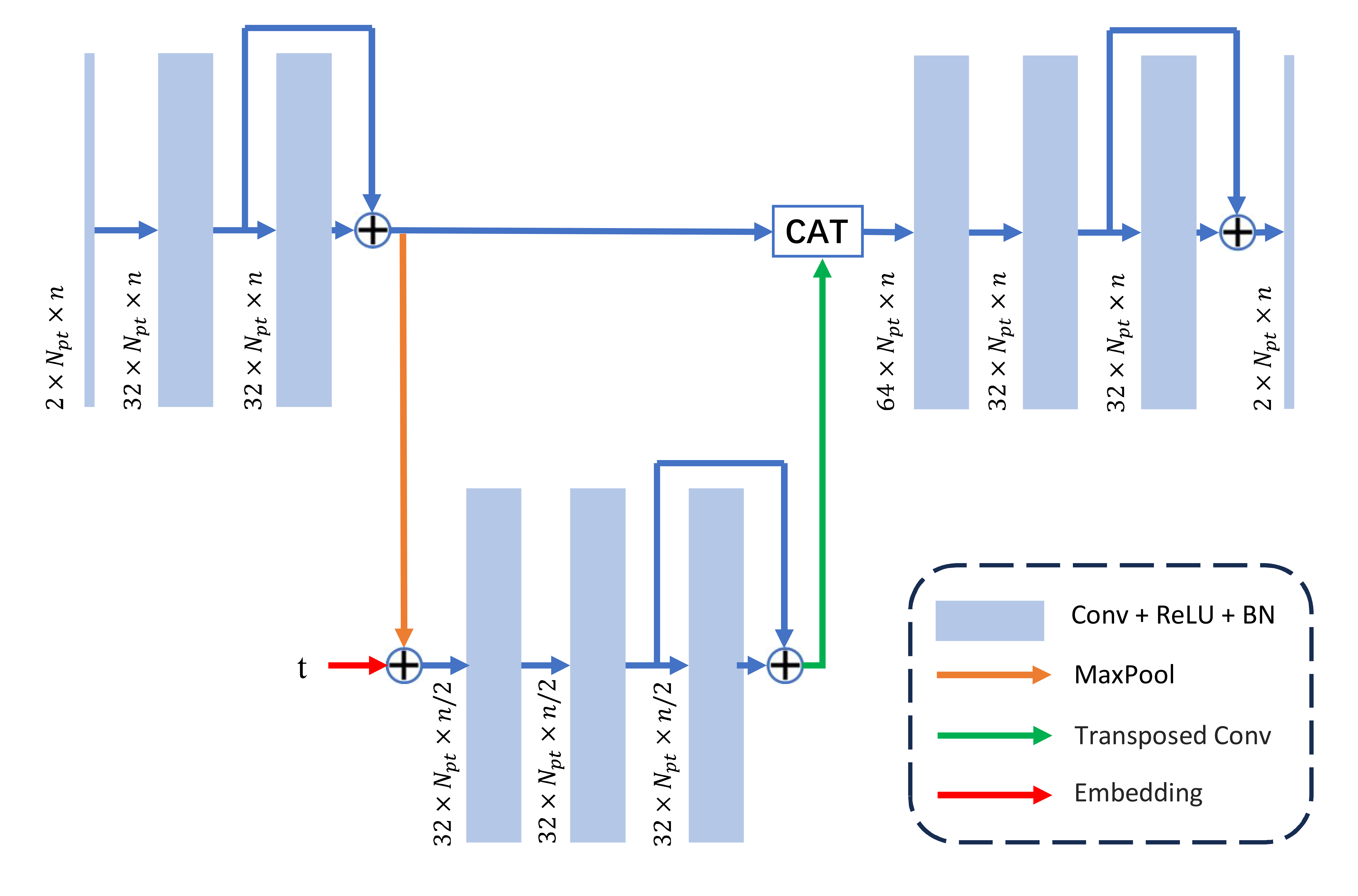}\\
		\caption{The architecture of U-Net in DMO. The symbol CAT denotes the concatenation of the given inputs in the channel dimension. \textcolor{black}{A  ``+'' in a circle indicates the direct summation of the given inputs.} 
		}
		\label{fig_NetArch}
	\end{figure}

	Given that most applications \textcolor{black}{of diffusion} models for image synthesis have favored U-net \cite{Unet} due to its strong ability to perform image-to-image translation, we use a U-Net architecture, as illustrated in Fig. \ref{fig_NetArch}. 
	The convolutional layer is integrated with the ReLU function and batch normalization to compose a block, which is the backbone of the \textcolor{black}{network}. However, a schedule differs from a typical image in that it lacks locality and spatial invariance, which are the inductive biases of convolutional neural networks. 
	\textcolor{black}{To accommodate this difference, the size of the convolution kernel is increased, ensuring that the schedule images remain within the receptive field of the convolutional neural network.} The convolution kernel size is configured as $(N_{pt}, 5)$, matching the problem's scale with the receptive field size. Furthermore, jump connections are incorporated between blocks to increase precision \cite{Resnet}. 
	Given the size of the schedule Gantt chart, the network includes only one layer each for up-sampling and down-sampling. Down-sampling is achieved using max pooling, while up-sampling utilizes transposed convolution.
	As $N_{pt}$ is small, it is not involved in up-sampling and down-sampling. The kernel size for both operations is fixed at $(1,2)$. 
	
	In practice, the input and output consist of schedules for a single component tank. To create a complete gasoline blending schedule, $N_{ct}$ such schedules are processed simultaneously. Given that convolutional neural networks are not sensitive to image size, DMO is expected to perform well on large-scale problems. 
	
	\subsection{Optimization in DMO}
	
	Using the diffusion model alone can only generate random schedules, whereas our objective is to generate Pareto-optimal schedules. For image synthesis, diffusion models generate the desired content by iteratively adding the gradient of a discriminator to the image: 
	\begin{equation}
		\begin{aligned}
			x_{t-1} &\sim \mathcal{N} (\tilde{\bf\mu}(x_t,\epsilon) + s\nabla_{x_t}\text{Loss}(f_\phi(x_t), y), \tilde{\beta}_t\bf{I}) \\
		\end{aligned}
		\label{eq3.6}
	\end{equation}
	where $s$ represents the  gradient scale, $y$ denotes the desired label or regression value, and $f_\phi$ is \textcolor{black}{a discriminator that has been well-trained} on noisy data. 
	
	For the gasoline blending scheduling problem, DMO aims to generate schedule images with optimal objectives. Drawing inspiration from the strategy used to generate specific content images, the discriminator is replaced by the objective functions. The objectives $E_{blend}$ and $E_{yield}$ are simultaneously optimized directly using the gradient descent method during the iteration of the diffusion model. However, not all constraints can be resolved using diffusion models, such as capacity constraints. These constraints are converted into the objective $E_{const}$, which is greater than 0 only when the constraints are violated. 
	To obtain the Pareto front, a series of solutions $X_t=(x^1_t,...,x^N_t)$ needs to be generated, so different objective weights \textbf{w} are assigned to each solution: 
	\begin{equation}
		\begin{aligned}
			&X_{t-1} \sim \mathcal{N}(\mu_\theta(X_t) + s\cdot g, \Sigma_\theta(X_t)) \\
			\text{where } &g = \nabla_{X_t}[\textbf{w}\cdot (E_{blend}, E_{yield})+ E_{const}]
		\end{aligned}
		\label{eq3.7}
	\end{equation} 
	
	Upon completion of the diffusion model, $X_1$ needs to be standardized to address minor constraint violations. In the standardization step, the parts of the solutions that violate the constraints are scaled down to ensure that the solutions strictly conform to the constraints. 
	Subsequently, \textcolor{black}{nondominanted} solutions can be derived from $X_1$, \textcolor{black}{which represents} the Pareto front. 
	
	In a typical optimization approach, the common practice is to first ensure that the solutions adhere to the constraints and then optimize their objectives. For instance, the constraint preference rule \cite{Debc} for addressing constraints in genetic algorithms involves favoring solutions in the population that satisfy the constraints. However, this approach has two drawbacks: 
	
	\begin{itemize}
		\item [1)]
		During the process of optimizing the objectives, some solutions may be eliminated due to minor constraint violations, leading to a significant waste of iterations. In extreme cases, the algorithm may fail to find a feasible solution even after numerous iterations. 
		\item [2)]
		The initial batch of constraint-compliant solutions generated heavily influences the population. Consequently, the performance of typical methods tends to be unstable. 
	\end{itemize}
	\textcolor{black}{In contrast, DMO tackles optimality and feasibility simultaneously. }DMO optimizes the objectives while ensuring that \textcolor{black}{the} solutions adhere to the constraints by creating multiple intermediate distributions between the Gaussian noise and the feasible domain. This strategy enables DMO to explore the solution space more efficiently. 
	
	\section{Simulation Studies}
	
	\subsection{Case Introduction}
	
	To empirically evaluate the performance of the proposed DMO method, a series of experiments was conducted using historical supply and demand data from a real-world refinery located in mainland China. 
	The blending process involved five blending components: reformed gasoline, non-aromatic gasoline, catalytic gasoline, MTBE, and C5. The properties of the gasoline considered included \textcolor{black}{the} octane number (ON), Reid vapor pressure (RVP), and sulfur content. The refinery produced three types of gasoline: regular, plus, and premium, each with different ON and RVP values. Therefore, the number of component and product tanks was set to (5, 3). To test the algorithm's performance on larger-scale problems, two additional cases were designed based on the available data, each with a different number of component and product tanks: (8, 5) and (12, 7). 
	The scheduling time is adjusted to 20, 100, and 300 time units, corresponding to two days, 10 days, and one month, respectively. These scenarios, which involved up to 50,400 decision variables (the dimension is calculated by $2\times N_{ct}\times N_{pt}\times n $), provide a robust testing ground to evaluate \textcolor{black}{the ability of DMO} to solve large-scale scheduling optimization problems. 
	
	\textcolor{black}{The data used to train the diffusion model were also from the historical data of this refinery. A year's worth of data on specific operations related to gasoline blending scheduling at this refinery were collected. A scheduling Gantt chart spanning more than 4,000 time units was obtained by processing these data. For training the model, specific time segments and component tank data were directly extracted from the Gantt chart. }
	
	\subsection{Experimental Settings}
	
	In this study, five representative MOEAs are compared: NSGA-II \cite{NSGA-II}, WOF-NSGA-II \cite{WOF}, MOEA/D-LWS \cite{MOEA/D-LWS}, MOEA/HD \cite{MOEA/HD}, and GMOEA \cite{GMOEA}. 
	NSGA-II was selected as the most widely used MOEA in the field. WOF-NSGA-II was chosen as a large-scale MOEA, specifically designed to handle problems with a large number of decision variables. MOEA/D-LWS and MOEA/HD are advanced MOEAs based on decomposition techniques. GMOEA is a MOEA that incorporates deep learning methods into its optimization process. 
	
	For fair \textcolor{black}{comparison}, the algorithms \textcolor{black}{were} enhanced based on the characteristics of the gasoline blending scheduling problem. The specific methods of improvement \textcolor{black}{were} as follows: 
	
	\begin{itemize}
		\item [1)]\textit{NSGA-II}: 
		In addition to polynomial mutation \cite{PM}, an operator that performs recombination based on time slices for offspring generation \textcolor{black}{was} also adopted \cite{EA1}.  
		\item [2)]\textit{WOF-NSGA-II}: 
		The WOF framework does not include a new or enhanced grouping mechanism. Instead, it utilizes a generic grouping mechanism that can incorporate any grouping mechanism from the literature \cite{WOF}. Based on the problem characteristics, the time \textcolor{black}{was} evenly divided as a grouping method. This approach \textcolor{black}{ensured} that solutions satisfying the constraints of the subproblems also satisfy the constraints of the overall problem. 
		\item [3)]\textit{\textcolor{black}{MOEA/D-LWS}}: 
		\textcolor{black}{The \textcolor{black}{distributions} of the decomposition weights in the algorithm \textcolor{black}{were} adjusted according to the scales of the two objectives. Additionally, an operator that conducts recombination using time slices \textcolor{black}{was} incorporated, similar to NSGA-II. }
		\item [4)]\textit{\textcolor{black}{MOEA/HD}}: 
		\textcolor{black}{While MOEA/HD uses a dynamic, hierarchical decomposition approach theoretically designed to accommodate non-standard Pareto fronts, the objectives are still normalized to optimize the algorithm's performance. }
		\item [5)]\textit{GMOEA}: 
		The discriminator in GMOEA is responsible for differentiating the optimal solutions. To improve the algorithm, a multiple discriminator approach was incorporated, which is a common improvement technique in GANs \cite{DDcGAN}. In this case, an additional discriminator \textcolor{black}{was} designed for GMOEA to assess whether a solution \textcolor{black}{adhered} to the imposed constraints. This enables the generation of a greater number of new solutions that \textcolor{black}{would} satisfy the given constraints.
	\end{itemize}
	
	These improvements \textcolor{black}{aimed to} significantly enhance the efficiency of these algorithms for the gasoline blending scheduling problem. In many cases, algorithms without these enhancements may struggle to find feasible solutions even with a large population size and number of iterations. 
	
	\textcolor{black}{NSGA-II, WOF-NSGA-II, MOEA/D-LWS, and MOEA/HD \textcolor{black}{were} implemented in Geatpy \cite{geatpy} using Python 3.10.9. GMOEA \textcolor{black}{used} to its official implementation \cite{GMOEA}.} DMO was implemented in PyTorch 1.13 using Python 3.10.9. All the algorithms \textcolor{black}{were} executed on a PC equipped with an Intel Core i9-10900K 3.7-GHz processor, 64 GB of RAM, and an Nvidia GeForce RTX 4090 GPU. 
	
	\begin{itemize}
		\item [1)]\textit{Population Size}: 
		The population size \textcolor{black}{was} set to 1024. This choice is based on the fact that DMO primarily utilizes GPU calculations, and \textcolor{black}{setting the population size as a power of 2 is often beneficial for efficient GPU processing.} 
		\item [2)]\textit{Termination Condition}: 
		The maximum number of iterations of NSGA-II, WOF-NSGA-II, MOEA/D-LWS, MOEA/HD, and GMOEA \textcolor{black}{was} set to 500. However, for DMO, the number of iterations needs to be determined before training. Therefore, three sets of models \textcolor{black}{were trained with 200, 500, and 1000 iterations. }
		\item [3)]\textit{Training Parameters}: 
		In GMOEA, the training parameters of the GANs \textcolor{black}{were} set according to \textcolor{black}{the} original paper. The additional discriminator \textcolor{black}{was} pre-trained, and its parameters remained unchanged during the GMOEA run. 
		In our proposed DMO, during the training process, we set the batch size to 4096 and the initial learning rate to $10^{-3}$, and \textcolor{black}{we conducted training for 1000 epochs.} This training phase typically requires approximately three hours when executed on an RTX 4090 GPU. Once the loss function \textcolor{black}{stabilized} without further changes, we \textcolor{black}{considered} the training complete. \textcolor{black}{The optimizer used in model training was the Adam optimizer, with $\beta_1=0.9$, $\beta_2=0.999$, and $\epsilon=10^{-8}$. The learning rate scheduler \textcolor{black}{utilized} a warm-up strategy comprising 20 warm-up steps. During the optimization process, the value of $s$ \textcolor{black}{was} fixed at $0.05$. Additionally, the weights of the objectives, denoted by \textbf{w}, \textcolor{black}{were} uniformly assigned in the ranges $(0.3, 0.7)$ and $(0.7, 0.3)$. This choice reflects our preference for solutions that avoid extremes. }
	\end{itemize}
	
	Afterward, each algorithm \textcolor{black}{was} independently executed 20 times on the gasoline blending scheduling problem, using different parameters. Subsequently, the results obtained by the proposed GMOEA and the compared algorithms \textcolor{black}{were} compared using the Wilcoxon rank-sum test [33] at a significance level of 0.05. The symbols ``$+$", ``$-$", and ``$\approx$" indicate \textcolor{black}{that} the compared algorithm performs significantly better than, significantly worse than, and statistically \textcolor{black}{the same as} the chosen competitor, respectively. 
	
	\subsection{Performance Indicators}
	
	\begin{table*}[hbp]
		\arrayrulecolor{black}
		\centering
		\begin{threeparttable}
			\color{black}\caption{HV Results Obtained by NSGA-II, WOF-NSGA-II, MOEA/D-LWS, MOEA/HD, GMOEA, and DMO ($T=200/500/1000$) on the Gasoline Blending Scheduling Problem. The Best Result in Each Row is in Boldface. The Symbols ``$+$", ``$-$", and ``$\approx$" Indicate that the Compared Algorithm is Significantly Better than, Significantly Worse than, and Statistically Tied with the Best Algotithm, Respectively. }
			\label{tab_HV}
			\begin{tabular}{cccccc}
				\toprule
				Scale & $n$ (Time) & NSGA-II & WOF-NSGA-II & MOEA/D-LWS & MOEA/HD \\
				\midrule
				\multirow{3}[1]{*}{\thead{$N_{ct}=5$\\$N_{pt}=3$}}
				& 20  & 0.574(1.917e-1)$-$ & 0.529(1.922e-1)$-$ & 0.405(2.507e-1)$-$ & 0.463(2.463e-1)$-$ \\
				& 100 & 0.289(2.309e-1)$-$ & 0.464(1.989e-1)$-$ & 0.000(0.000e-0)$-$ & 0.046(6.934e-3)$-$ \\
				& 300 & - & 0.412(1.935e-1)$-$ & - & - \\
				\midrule
				\multirow{3}[1]{*}{\thead{$N_{ct}=8$\\$N_{pt}=5$}}
				& 20  & 0.457(2.201e-1)$-$ & 0.467(1.913e-1)$-$ & 0.125(5.788e-2)$-$ & 0.242(6.578e-2)$-$ \\
				& 100 & - & 0.390(1.920e-1)$-$ & - & - \\
				& 300 & - & 0.389(1.935e-1)$-$ & - & - \\
				\midrule
				\multirow{3}[1]{*}{\thead{$N_{ct}=12$\\$N_{pt}=7$}}
				& 20  & 0.207(1.964e-1)$-$ & 0.342(1.882e-1)$-$ & 0.000(0.000e-0)$-$ & 0.000(0.000e-0)$-$ \\
				& 100 & - & 0.330(1.902e-1)$-$ & - & - \\
				& 300 & - & 0.329(1.838e-1)$-$ & - & - \\
				\midrule
				\multicolumn{2}{c}{$+/-/\approx$} & $0/4/0$ & $0/9/0$ & $0/4/0$ & $0/4/0$ \\
				\toprule
				Scale & $n$ (Time) & GMOEA & DMO ($T=200$) & DMO ($T=500$) & DMO ($T=1000$) \\
				\midrule
				\multirow{3}[1]{*}{\thead{$N_{ct}=5$\\$N_{pt}=3$}}
				& 20  & 0.581(2.122e-2)$-$ & 0.965(1.580e-5)$-$ & 0.979(1.124e-5)$\approx$  & \textbf{0.990(2.157e-5)} \\
				& 100 & 0.478(4.489e-2)$-$ & 0.950(3.164e-5)$-$ & 0.971(2.338e-5)$-$  & \textbf{0.987(1.743e-5)} \\
				& 300 & 0.402(5.019e-2)$-$ & 0.949(5.219e-5)$-$ & 0.972(3.018e-5)$-$ & \textbf{0.983(2.668e-5)} \\
				\midrule
				\multirow{3}[1]{*}{\thead{$N_{ct}=8$\\$N_{pt}=5$}}
				& 20  & 0.508(5.502e-2)$-$ & 0.826(4.282e-5)$-$ & 0.858(2.322e-5)$-$ & \textbf{0.901(1.559e-5)} \\
				& 100 & 0.418(4.523e-2)$-$ & 0.769(6.424e-5)$-$ & 0.842(2.993e-5)$-$ & \textbf{0.883(2.187e-5)} \\
				& 300 & 0.348(4.201e-2)$-$ & 0.785(1.285e-4)$-$ & 0.833(4.981e-5)$-$ & \textbf{0.891(3.120e-5)} \\
				\midrule
				\multirow{3}[1]{*}{\thead{$N_{ct}=12$\\$N_{pt}=7$}}
				& 20  & 0.451(5.770e-2)$-$ & 0.659(9.968e-5)$-$ & 0.743(2.677e-5)$-$ & \textbf{0.843(1.967e-5)} \\
				& 100 & 0.381(7.774e-2)$-$ & 0.667(1.917e-4)$-$ & 0.759(3.753e-5)$-$ & \textbf{0.830(3.152e-5)} \\
				& 300 & 0.308(6.906e-2)$-$ & 0.629(3.191e-4)$-$ & 0.726(5.889e-5)$-$ & \textbf{0.843(4.132e-5)} \\
				\midrule
				\multicolumn{2}{c}{$+/-/\approx$} & $0/9/0$ & $0/9/0$ & $0/8/1$ &  \\
				\bottomrule
			\end{tabular}
			\begin{tablenotes}
				\item[1] The ``-'' symbol indicates that the algorithm is unable to generate feasible solutions for the corresponding case of the gasoline blending scheduling problem. 
			\end{tablenotes}
		\end{threeparttable}
	\end{table*}
	
	\begin{table*}[htbp]
		\arrayrulecolor{black}
		\centering
		\begin{threeparttable}[b]
			\color{black}\caption{C-metric Results between DMO(A) ($T=200/500/1000$) and NSGA-II / WOF-NSGA-II / MOEA/D-LWS / MOEA/HD / GMOEA(B) on the Gasoline Blending Scheduling Problem. The Symbols ``$+$", ``$-$", and ``$\approx$" Indicate that $C(B,A)$ is Significantly Greater than,  Significantly Less than, and Statistically Tied with $C(A,B)$.}
			\label{tab_CM}
			\begin{tabular}{ccccccccc}
				\toprule
				DMO$(A)$ & Scale & $n$ (Time) &  & NSGA-II$(B)$ & WOF-NSGA-II$(B)$ & MOEA/D-LWS$(B)$ & MOEA/HD$(B)$ & GMOEA$(B)$ \\
				\midrule
				\multirow{18}[1]{*}{$T=200$}
				& \multirow{6}[1]{*}{\thead{$N_{ct}=5$\\$N_{pt}=3$}}  
				& \multirow{2}[1]{*}{20}
				& $C(A,B)$ & \textbf{0.475(8.309e-2)} & \textbf{0.460(9.062e-2)} & \textbf{0.537(2.538e-2)} & \textbf{0.505(2.179e-2)} & \textbf{0.422(7.653e-3)}\\
				&&& $C(B,A)$& 0.130(1.550e-2) & 0.275(1.594e-2) & 0.108(1.275e-2) & 0.108(1.365e-2) & 0.316(6.504e-3)\\
				\cmidrule{3-9}
				&& \multirow{2}[1]{*}{100}
				& $C(A,B)$ & \textbf{0.834(7.313e-2)} & \textbf{0.462(6.914e-2)} & \textbf{0.996(1.271e-2)} & \textbf{0.967(1.532e-2)} & \textbf{0.552(8.410e-3)}\\
				&&& $C(B,A)$& 0.049(9.506e-3) & 0.211(1.776e-2) & 0.000(0.000e-0) & 0.019(2.721e-3) & 0.200(7.472e-3)\\
				\cmidrule{3-9}
				&& \multirow{2}[1]{*}{300}
				& $C(A,B)$ & - & \textbf{0.470(7.893e-2)} & - & - & \textbf{0.757(8.836e-3)}\\
				&&& $C(B,A)$& - & 0.184(6.829e-3) & - & - & 0.154(1.275e-3)\\
				\cmidrule{2-9}
				& \multirow{6}[1]{*}{\thead{$N_{ct}=8$\\$N_{pt}=5$}}  
				& \multirow{2}[1]{*}{20}
				& $C(A,B)$ & \textbf{0.662(8.026e-2)} & \textbf{0.603(7.077e-2)} & \textbf{0.961(3.290e-2)} & \textbf{0.949(7.831e-2)} & \textbf{0.516(7.855e-3)}\\
				&&& $C(B,A)$& 0.111(1.480e-2) & 0.150(1.175e-2) & 0.013(2.347e-3) & 0.016(3.149e-3) & 0.310(1.612e-2)\\
				\cmidrule{3-9}
				&& \multirow{2}[1]{*}{100}
				& $C(A,B)$ & - & \textbf{0.615(7.455e-2)} & - & - & \textbf{0.626(8.450e-3)}\\
				&&& $C(B,A)$& - & 0.141(1.130e-2) & - & - & 0.297(1.593e-2)\\
				\cmidrule{3-9}
				&& \multirow{2}[1]{*}{300}
				& $C(A,B)$ & - & \textbf{0.668(6.977e-2}) & - & - & \textbf{0.781(7.636e-3)}\\
				&&& $C(B,A)$& - & 0.103(9.502e-3) & - & - & 0.137(2.641e-3)\\
				\cmidrule{2-9}
				& \multirow{6}[1]{*}{\thead{$N_{ct}=12$\\$N_{pt}=7$}}  
				& \multirow{2}[1]{*}{20}
				& $C(A,B)$ & \textbf{0.790(7.280e-2)} & \textbf{0.723(6.482e-2)} & \textbf{1.000(0.000e-0)} & \textbf{1.000(0.000e-0)} & \textbf{0.667(8.050e-3)}\\
				&&& $C(B,A)$& 0.109(7.857e-3) & 0.111(6.075e-3) & 0.000(0.000e-0) & 0.000(0.000e-0) & 0.279(6.694e-3)\\
				\cmidrule{3-9}
				&& \multirow{2}[1]{*}{100}
				& $C(A,B)$ & - & \textbf{0.783(7.078e-2)} & - & - & \textbf{0.802(1.590e-3)}\\
				&&& $C(B,A)$& - & 0.064(2.932e-3) & - & - & 0.158(2.295e-3)\\
				\cmidrule{3-9}
				&& \multirow{2}[1]{*}{300}
				& $C(A,B)$ & - & \textbf{0.823(5.074e-2)} & - & - & \textbf{0.897(7.649e-3)}\\
				&&& $C(B,A)$& - & 0.052(2.212e-3) & - & - & 0.065(3.157e-4)\\
				\midrule
				\multirow{18}[1]{*}{$T=500$}
				& \multirow{6}[1]{*}{\thead{$N_{ct}=5$\\$N_{pt}=3$}}  
				& \multirow{2}[1]{*}{20}
				& $C(A,B)$ & \textbf{0.603(6.120e-2)} & \textbf{0.590(6.454e-2)} & \textbf{0.658(5.930e-2)} & \textbf{0.566(5.184e-2)} & \textbf{0.541(6.177e-3)}\\
				&&& $C(B,A)$& 0.120(1.004e-1) & 0.172(1.154e-2) & 0.096(1.588e-2) & 0.122(1.248e-2) & 0.368(9.930e-3)\\
				\cmidrule{3-9}
				&& \multirow{2}[1]{*}{100}
				& $C(A,B)$ & \textbf{0.953(8.503e-2)} & \textbf{0.650(5.656e-2)} & \textbf{1.000(0.000e-0)} & \textbf{1.000(0.000e-0)} & \textbf{0.635(9.497e-3)}\\
				&&& $C(B,A)$& 0.016(6.213e-3) & 0.168(1.721e-2) & 0.000(0.000e-0) & 0.000(0.000e-0) & 0.189(5.502e-3)\\
				\cmidrule{3-9}
				&& \multirow{2}[1]{*}{300}
				& $C(A,B)$ & - & \textbf{0.694(9.389e-2)} & - & - & \textbf{0.759(4.479e-3)}\\
				&&& $C(B,A)$& - & 0.121(8.059e-3) & - & - & 0.092(2.544e-3)\\
				\cmidrule{2-9}
				& \multirow{6}[1]{*}{\thead{$N_{ct}=8$\\$N_{pt}=5$}}  
				& \multirow{2}[1]{*}{20}
				& $C(A,B)$ & \textbf{0.842(4.218e-2)} & \textbf{0.833(4.427e-2)} & \textbf{1.000(0.000e-0)} & \textbf{0.995(1.312e-2)} & \textbf{0.689(4.747e-3)}\\
				&&& $C(B,A)$& 0.048(2.043e-2) & 0.094(3.440e-3) & 0.000(0.000e-0) & 0.000(0.000e-0) & 0.219(2.043e-3)\\
				\cmidrule{3-9}
				&& \multirow{2}[1]{*}{100}
				& $C(A,B)$ & - & \textbf{0.864(7.279e-2)} & - & - & \textbf{0.904(9.369e-3)}\\
				&&& $C(B,A)$& - & 0.052(1.130e-3) & - & - & 0.091(1.048e-3)\\
				\cmidrule{3-9}
				&& \multirow{2}[1]{*}{300}
				& $C(A,B)$ & - & \textbf{0.901(6.374e-2)} & - & - & \textbf{0.945(2.957e-3)}\\
				&&& $C(B,A)$& - & 0.017(9.502e-4) & - & - & 0.047(8.168e-4)\\
				\cmidrule{2-9}
				& \multirow{6}[1]{*}{\thead{$N_{ct}=12$\\$N_{pt}=7$}}  
				& \multirow{2}[1]{*}{20}
				& $C(A,B)$ & \textbf{0.912(7.636e-3)} & \textbf{0.900(3.689e-3)} & \textbf{1.000(0.000e-0)} & \textbf{1.000(0.000e-0)} & \textbf{0.775(4.817e-3)}\\
				&&& $C(B,A)$& 0.021(1.369e-3) & 0.023(1.102e-3) & 0.000(0.000e-0) & 0.000(0.000e-0) & 0.121(2.464e-3)\\
				\cmidrule{3-9}
				&& \multirow{2}[1]{*}{100}
				& $C(A,B)$ & - & \textbf{0.913(3.266e-3)} & - & - & \textbf{0.821(4.076e-3)}\\
				&&& $C(B,A)$& - & 0.017(6.379e-4) & - & - & 0.064(8.531e-4)\\
				\cmidrule{3-9}
				&& \multirow{2}[1]{*}{300}
				& $C(A,B)$ & - & \textbf{0.917(5.481e-3)} & - & - & \textbf{0.947(6.621e-3)}\\
				&&& $C(B,A)$& - & 0.010(5.282e-4) & - & - & 0.030(3.234e-4)\\
				\midrule
				\multirow{18}[1]{*}{$T=1000$}
				& \multirow{6}[1]{*}{\thead{$N_{ct}=5$\\$N_{pt}=3$}}  
				& \multirow{2}[1]{*}{20}
				& $C(A,B)$ & \textbf{0.726(2.799e-2)} & \textbf{0.606(2.002e-2)} & \textbf{0.856(3.081e-2)} & \textbf{0.753(3.269e-2)} & \textbf{0.571(3.470e-3)}\\
				&&& $C(B,A)$& 0.113(4.446e-3) & 0.167(4.465e-3) & 0.081(8.406e-3) & 0.141(1.582e-2) & 0.212(7.760e-3)\\
				\cmidrule{3-9}
				&& \multirow{2}[1]{*}{100}
				& $C(A,B)$ & \textbf{0.978(1.447e-2)} & \textbf{0.733(2.627e-2)} & \textbf{1.000(0.000e-0)} & \textbf{1.000(0.000e-0)} & \textbf{0.672(2.580e-3)}\\
				&&& $C(B,A)$& 0.025(6.947e-4) & 0.142(3.435e-3) & 0.000(0.000e-0) & 0.000(0.000e-0) & 0.125(4.817e-3)\\
				\cmidrule{3-9}
				&& \multirow{2}[1]{*}{300}
				& $C(A,B)$ & - & \textbf{0.745(3.589e-2)} & - & - & \textbf{0.734(3.936e-3)}\\
				&&& $C(B,A)$& - & 0.104(1.489e-3) & - & - & 0.065(7.760e-4)\\
				\cmidrule{2-9}
				& \multirow{6}[1]{*}{\thead{$N_{ct}=8$\\$N_{pt}=5$}}  
				& \multirow{2}[1]{*}{20}
				& $C(A,B)$ & \textbf{0.903(1.106e-2)} & \textbf{0.881(3.725e-2)} & \textbf{1.000(0.000e-0)} & \textbf{1.000(0.000e-0)} & \textbf{0.697(2.779e-3)}\\
				&&& $C(B,A)$& 0.033(4.667e-3) & 0.031(2.018e-3) & 0.000(0.000e-0) & 0.000(0.000e-0) & 0.181(3.344e-3)\\
				\cmidrule{3-9}
				&& \multirow{2}[1]{*}{100}
				& $C(A,B)$ & - & \textbf{0.894(3.340e-2)} & - & - & \textbf{0.908(1.755e-3)}\\
				&&& $C(B,A)$& - & 0.029(1.422e-3) & - & - & 0.081(7.163e-4)\\
				\cmidrule{3-9}
				&& \multirow{2}[1]{*}{300}
				& $C(A,B)$ & - & \textbf{0.917(6.259e-2)} & - & - & \textbf{0.986(8.776e-3)}\\
				&&& $C(B,A)$& - & 0.011(4.664e-4) & - & - & 0.000(0.000e-0)\\
				\cmidrule{2-9}
				& \multirow{6}[1]{*}{\thead{$N_{ct}=12$\\$N_{pt}=7$}}  
				& \multirow{2}[1]{*}{20}
				& $C(A,B)$ & \textbf{0.980(1.011e-2)} & \textbf{0.906(9.681e-3)} & \textbf{1.000(0.000e-0)} & \textbf{1.000(0.000e-0)} & \textbf{0.828(3.717e-3)}\\
				&&& $C(B,A)$& 0.012(1.273e-3) & 0.013(1.435e-3) & 0.000(0.000e-0) & 0.000(0.000e-0) & 0.097(3.103e-3)\\
				\cmidrule{3-9}
				&& \multirow{2}[1]{*}{100}
				& $C(A,B)$ & - & \textbf{0.943(1.023e-2)} & - & - & \textbf{0.881(6.343e-3)}\\
				&&& $C(B,A)$& - & 0.010(3.435e-4) & - & - & 0.027(5.031e-4)\\
				\cmidrule{3-9}
				&& \multirow{2}[1]{*}{300}
				& $C(A,B)$ & - & \textbf{0.979(1.629e-2)} & - & - & \textbf{0.970(7.735e-3)}\\
				&&& $C(B,A)$& - & 0.010(7.244e-4) & - & - & 0.012(6.528e-4)\\
				\midrule
				\multicolumn{3}{c}{$+/-/\approx$} & & $0/12/0$ & $0/27/0$ & $0/12/0$ & $0/12/0$ & $0/27/0$  \\
				\bottomrule
			\end{tabular}
			\begin{tablenotes}
				\item[1] The ``-'' symbol indicates the absence of C-metric values for this case, as NSGA-II / MOEA/D-LWS / MOEA/HD fails to generate feasible solutions for the corresponding gasoline blending scheduling problem. 
			\end{tablenotes}
		\end{threeparttable}
	\end{table*}
	
	\begin{figure*}[htbp]
		\centering
		\subfigure[]{
			\begin{minipage}[b]{0.29\textwidth}
				\includegraphics[width=1\textwidth]{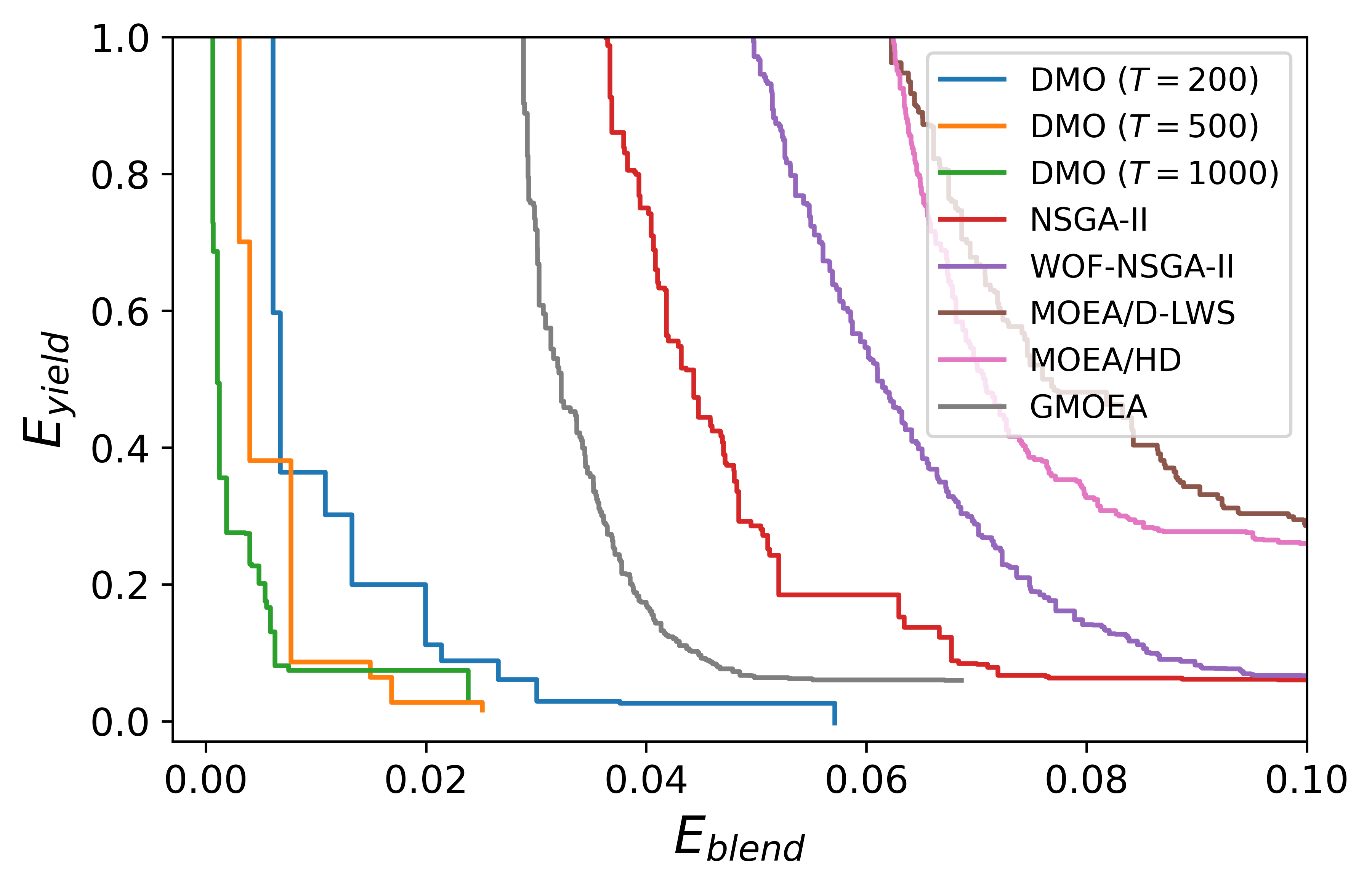}
			\end{minipage}\label{fig_ObjCompare_35}
		}
		\subfigure[]{
			\begin{minipage}[b]{0.29\textwidth}
				\includegraphics[width=1\textwidth]{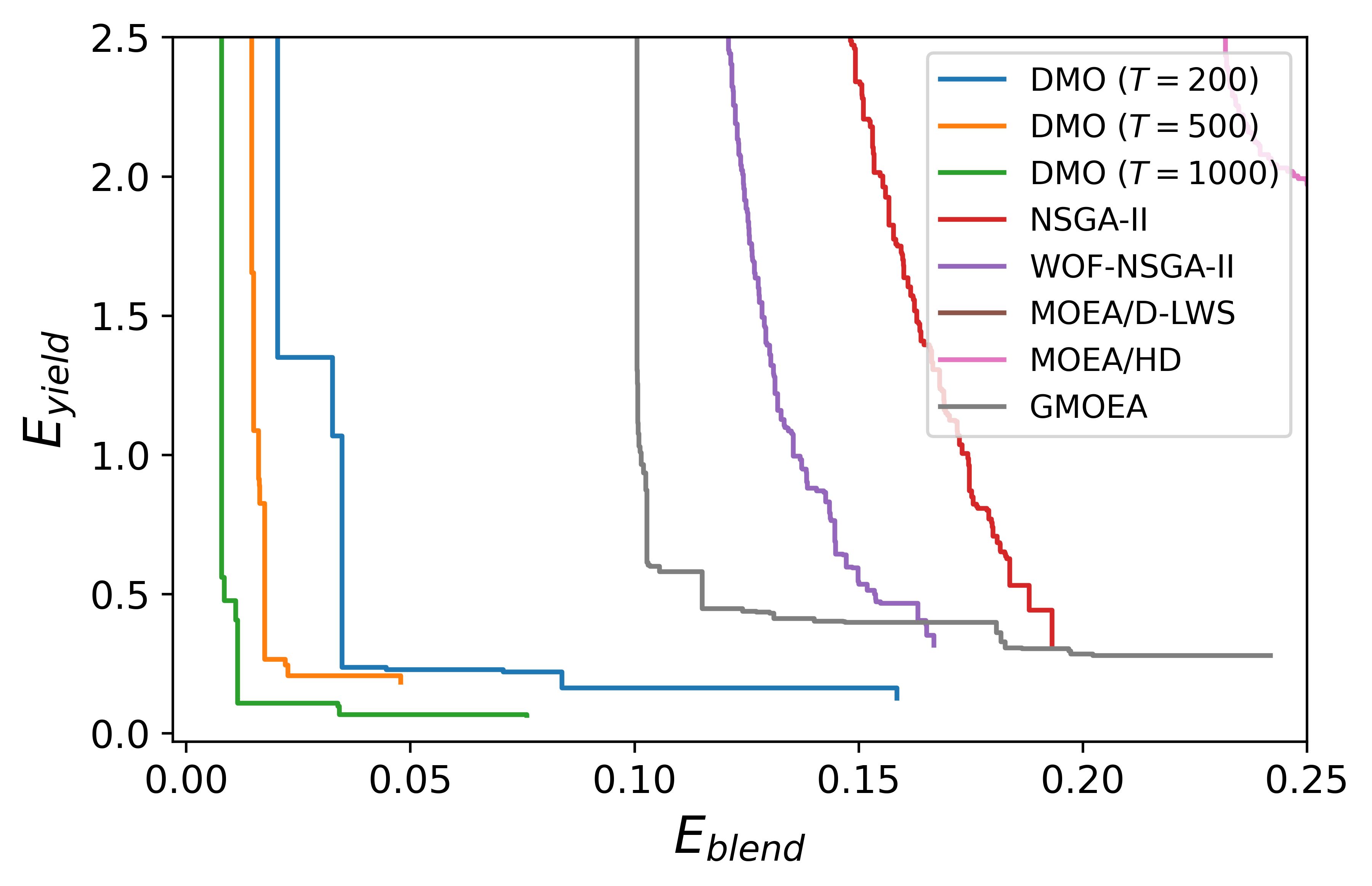}
			\end{minipage}\label{fig_ObjCompare_58}
		}
		\subfigure[]{
			\begin{minipage}[b]{0.29\textwidth}
				\includegraphics[width=1\textwidth]{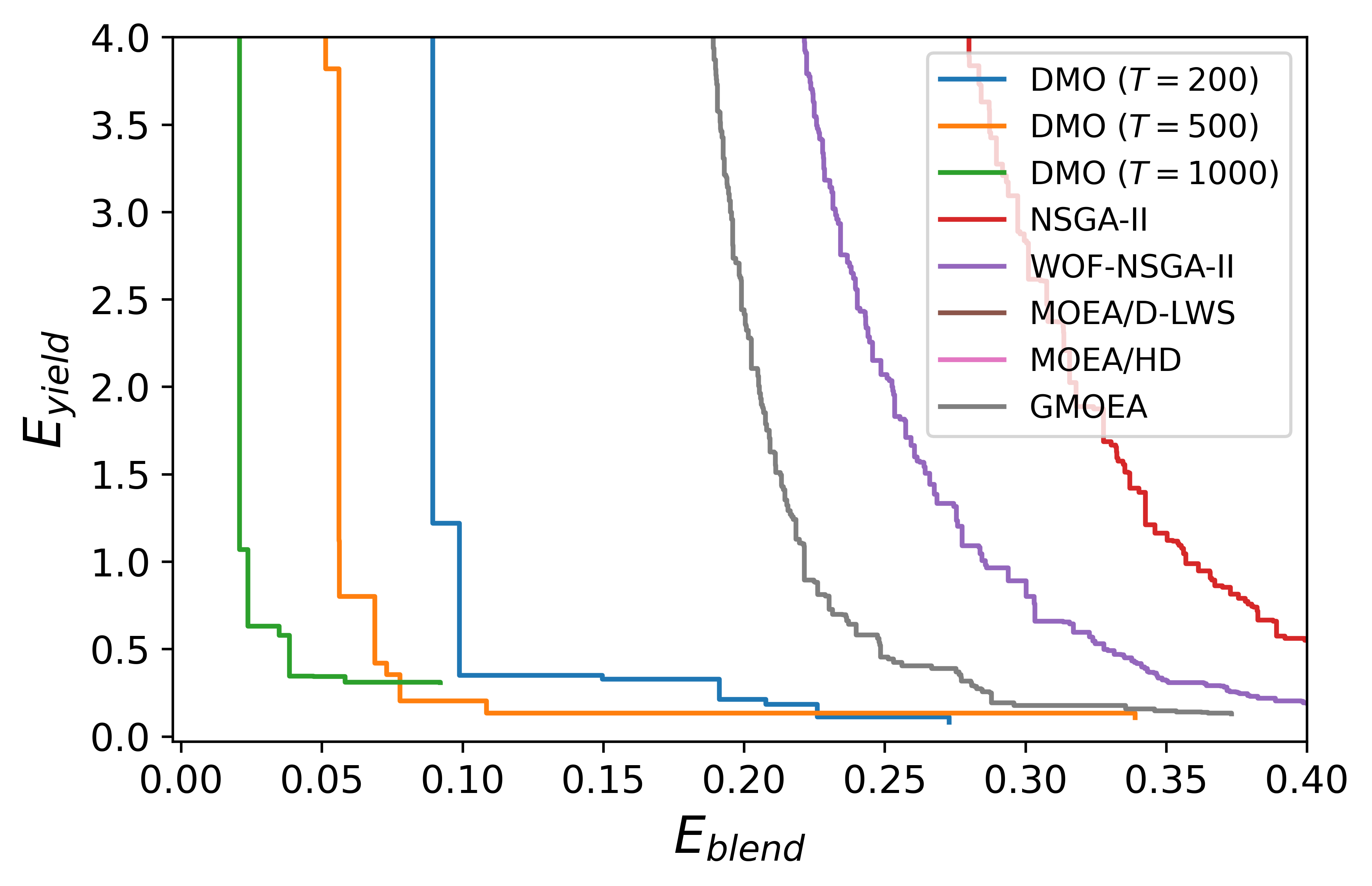}
			\end{minipage}\label{fig_ObjCompare_712}
		}
		\qquad
		\subfigure[]{
			\begin{minipage}[b]{0.29\textwidth}
				\includegraphics[width=1\textwidth]{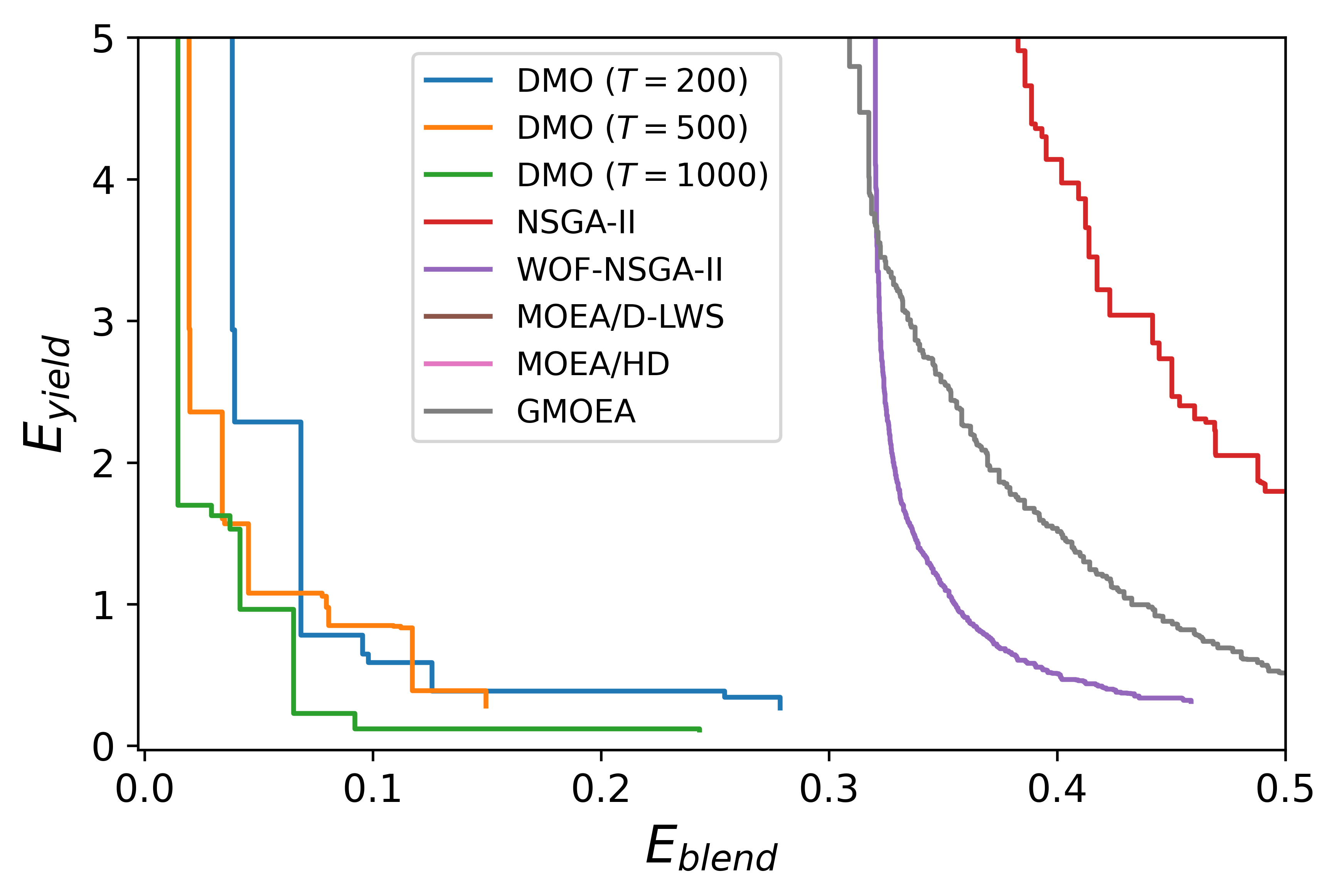}
			\end{minipage}\label{fig_ObjCompare_35_100}
		}
		\subfigure[]{
			\begin{minipage}[b]{0.29\textwidth}
				\includegraphics[width=1\textwidth]{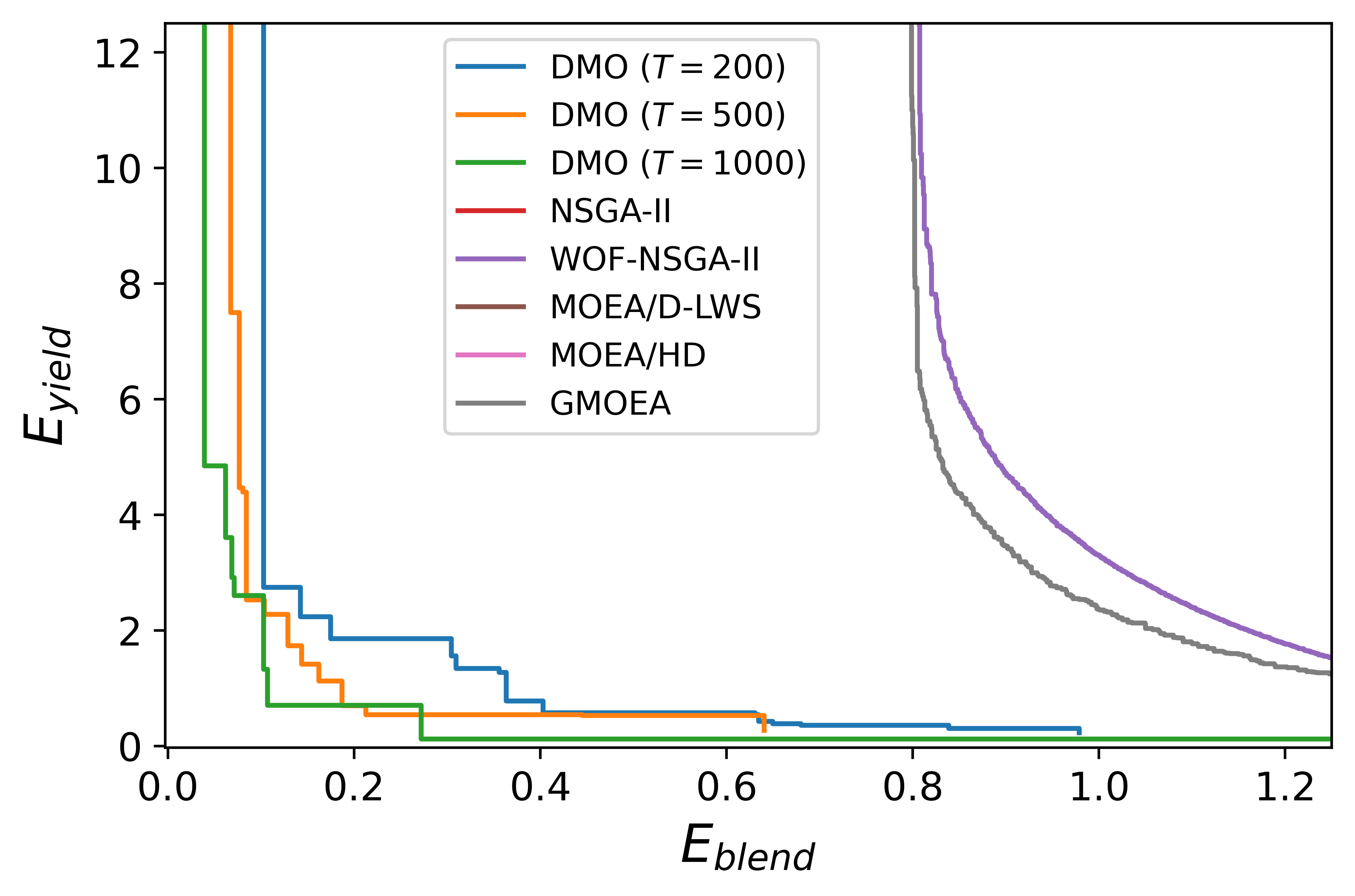}
			\end{minipage}\label{fig_ObjCompare_58_100}
		}
		\subfigure[]{
			\begin{minipage}[b]{0.29\textwidth}
				\includegraphics[width=1\textwidth]{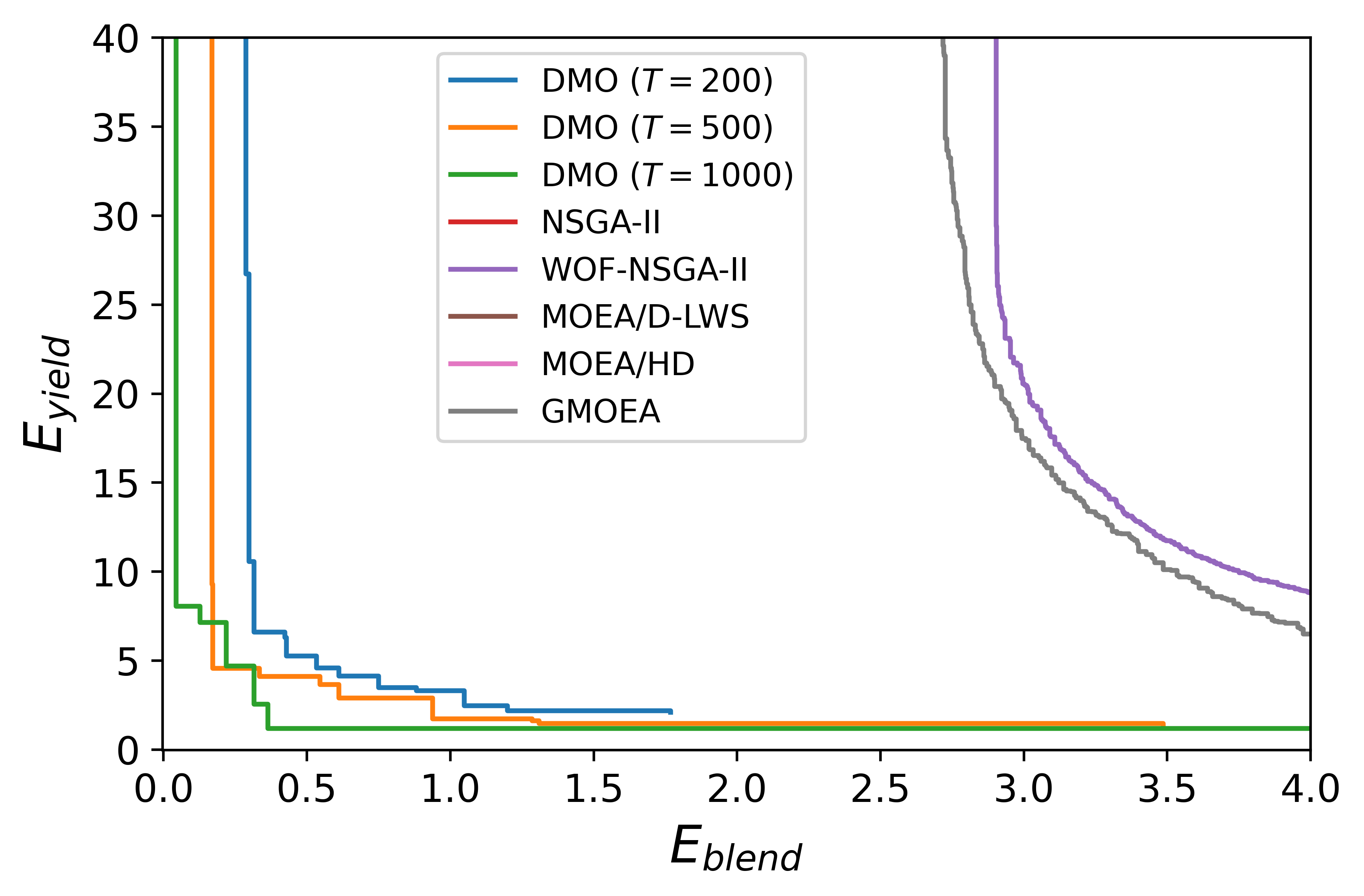}
			\end{minipage}\label{fig_ObjCompare_712_100}
		}
		\qquad
		\subfigure[]{
			\begin{minipage}[b]{0.29\textwidth}
				\includegraphics[width=1\textwidth]{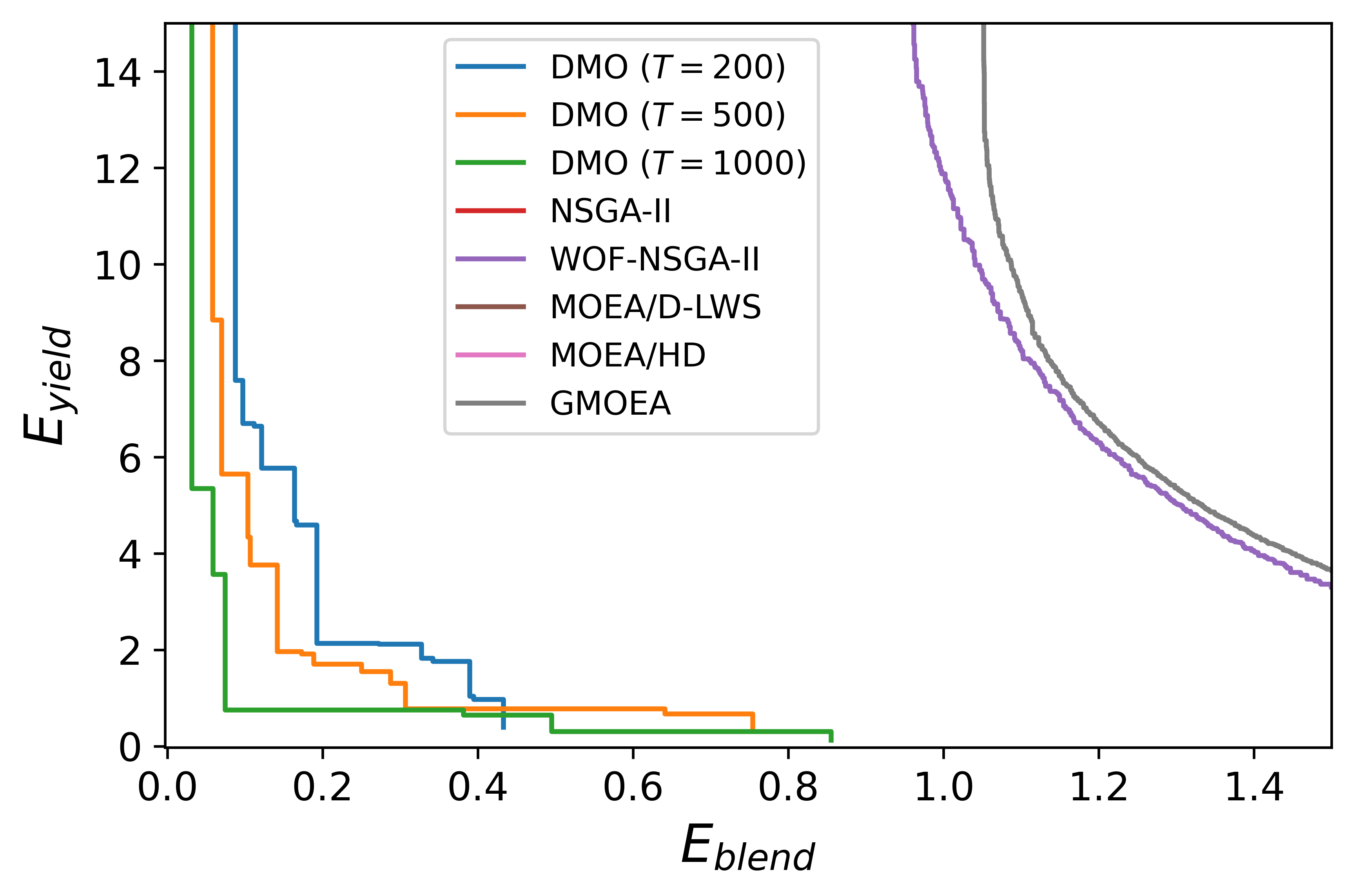}
			\end{minipage}\label{fig_ObjCompare_35_300}
		}
		\subfigure[]{
			\begin{minipage}[b]{0.29\textwidth}
				\includegraphics[width=1\textwidth]{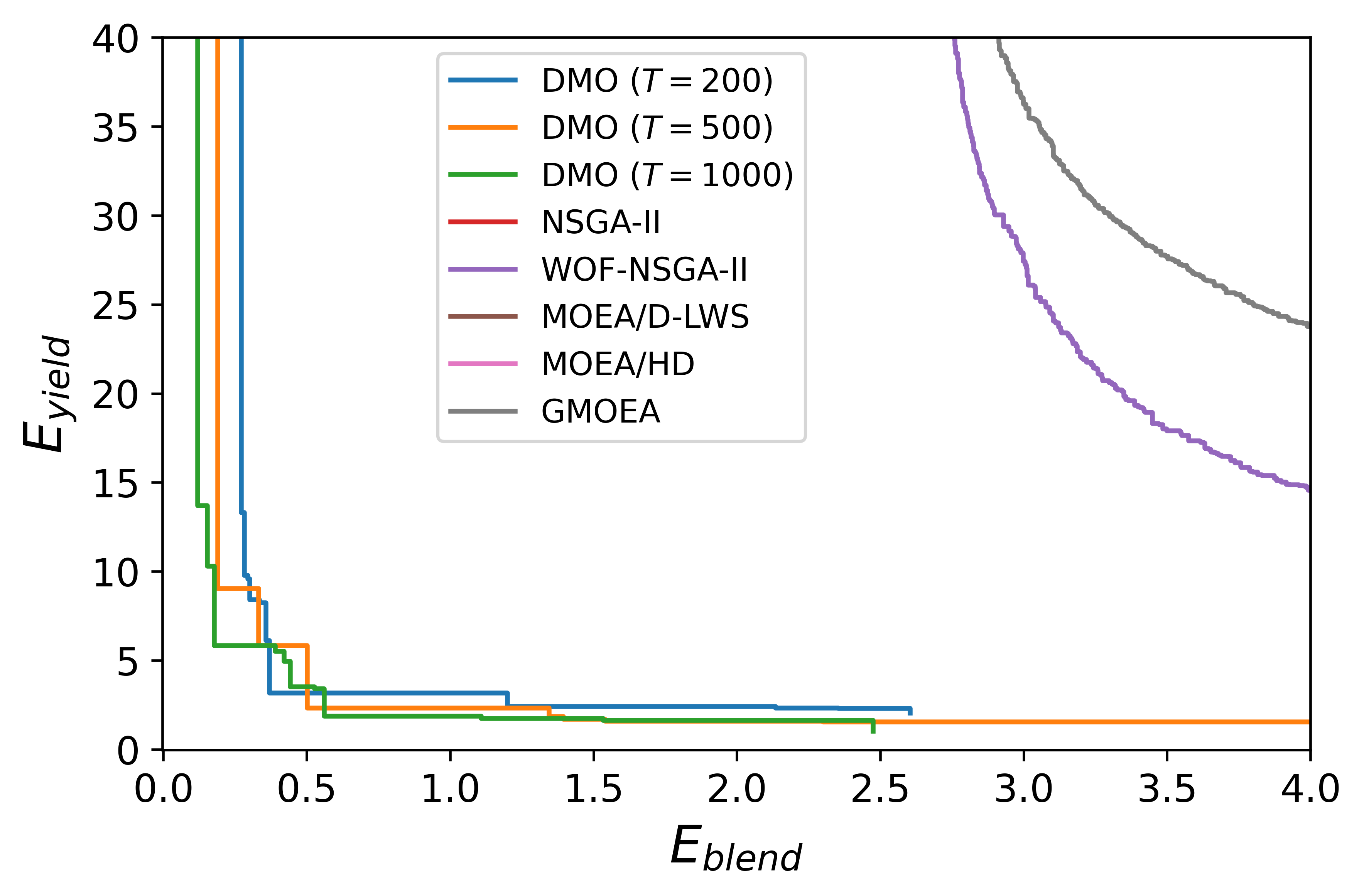}
			\end{minipage}\label{fig_ObjCompare_58_300}
		}
		\subfigure[]{
			\begin{minipage}[b]{0.29\textwidth}
				\includegraphics[width=1\textwidth]{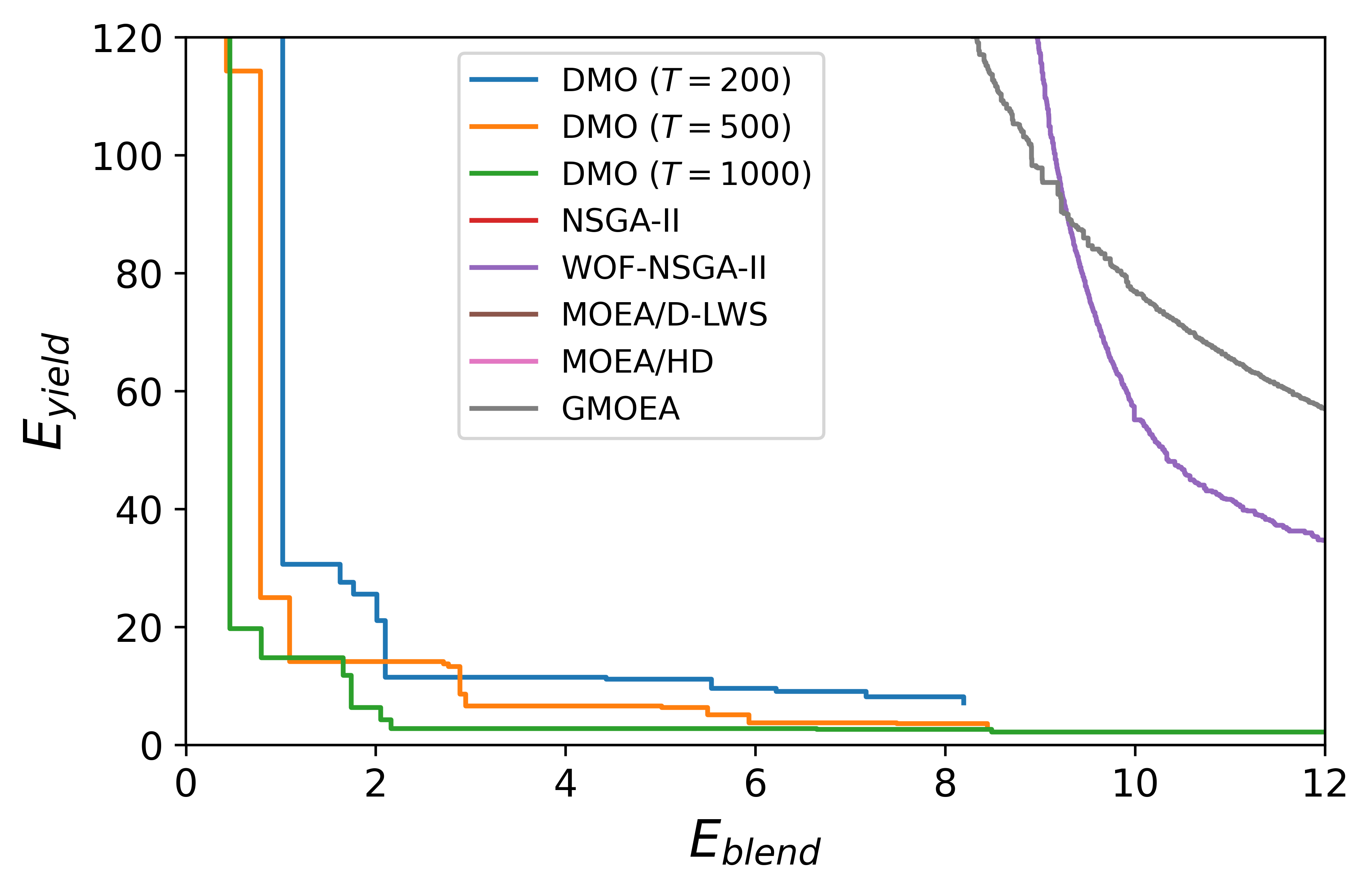}
			\end{minipage}\label{fig_ObjCompare_712_300}
		}
		\color{black}\caption{Median attainment surfaces obtained by DMO ($T=200/500/1000$), NSGA-II, WOF-NSGA-II, MOEA/D-LWS, MOEA/HD, and GMOEA on the gasoline blending scheduling problem. (a) $N_{ct}=5,N_{pt}=3,n=20$; (b) $N_{ct}=8,N_{pt}=5,n=20$; (c) $N_{ct}=12,N_{pt}=7,n=20$; (d) $N_{ct}=5,N_{pt}=3,n=100$; (e) $N_{ct}=8,N_{pt}=5,n=100$; (f) $N_{ct}=12,N_{pt}=7,n=100$; (g) $N_{ct}=5,N_{pt}=3,n=300$; (h) $N_{ct}=8,N_{pt}=5,n=300$; (i) $N_{ct}=12,N_{pt}=7,n=300$. } 
		\label{fig_ObjCompare}
	\end{figure*}
	
	\textcolor{black}{Two different performance indicators are adopted to assess the quality of the obtained results.} The first indicator is hypervolume (HV), which evaluates both the convergence and distribution of the obtained solution set without requiring knowledge of the true Pareto front \cite{HV}. Given a solution set $P$, the HV value of $P$ is defined as the area covered by $P$ with respect to a predefined reference point $\bf{r}$ in the objective space
	\begin{equation}
		\begin{aligned}
			HV(P,\boldsymbol{r})=\frac {\bigcup_{p\in P} hv(p,\boldsymbol{r})} {hv_O}
		\end{aligned}
		\label{eq3.8}
	\end{equation}
	where $hv(p,\boldsymbol{r})$ is the hypervolume enclosed \textcolor{black}{by} point $p$, and the reference points $\bf{r}$ and $hv_O$ represent the hypervolume enclosed between the origin and the reference point. A greater HV indicates better \textcolor{black}{algorithm performance}. In this case, the reference point is set to $ (0.005 \times n \times N_{pt}, 0.05 \times n \times N_{pt})$, where $n$ is the number of periods, and $N_{pt}$ is the number of product tanks. 
	
	The second performance indicator is the set coverage (C-metric) \cite{MOEAD}. The C-metric evaluates the convergence and distribution of \textcolor{black}{two algorithms} by comparing thier results. Let $A$ and $B$ represent two approximations to the Pareto front of a \textcolor{black}{multiobjective} problem. $C(A,B)$ is defined as the percentage of the solutions in $B$ that are dominated by at least one solution in $A$: 
	
	\begin{equation}
		\begin{aligned}
			C(A,B) = \frac{|\{b\in B|\exists a\in A: a \succ b\}|} {|B|}
		\end{aligned}
		\label{eq3.9}
	\end{equation}
	
	Note that $C(A,B)$ does not necessarily equal $1-C(B,A)$. $C(A,B)=1$ indicates that all the solutions in $B$ are dominated by at least one solution in $A$, while $C(A,B)=0$ implies that no solution in $B$ is dominated by any solution in $A$. 
	
	In addition to numerical indicators, we employ the median attainment surface to \textcolor{black}{assess} the results of each algorithm in the objective space. The median attainment surface is capable of \textcolor{black}{indicating} the actual performance of an algorithm with large variance \cite{MAS}. 
	
	\subsection{Performance of the proposed DMO}
	
	This section \textcolor{black}{compares the performance} of the proposed DMO and five other state-of-the-art MOEAs. The statistical results of the HV and \textcolor{black}{C-metrics} achieved by DMO and the five compared algorithms are summarized in Tables \ref{tab_HV} and \ref{tab_CM}, respectively. The median attainment surfaces obtained by DMO and the five compared algorithms are shown in Fig. \ref{fig_ObjCompare}. \textcolor{black}{From these results, the following conclusions can be drawn: }
	\begin{itemize}
		\item [1)]
		DMO exhibits superior overall performance when compared to the other algorithms in gasoline blending scheduling problems of any size. The advantage of DMO becomes more apparent as the problem size increases, indicating that DMO, which utilizes convolutional neural networks, is less affected by problem size. 
		\item [2)]
		The performance of DMO improves with \textcolor{black}{more} iterations of the diffusion model. Increased iterations provide DMO with greater optimization opportunities and enable it to effectively address conflicts between constraints and objectives.
		\item [3)]
		The advantages of DMO are more noticeable in terms of the HV metric than \textcolor{black}{in terms of} the C-metric. This is because the compared algorithms can generate extreme solutions that lie beyond the \textcolor{black}{area in which DMO's solutions dominate.} However, these solutions do not contribute to the HV metric \textcolor{black}{because} they lie beyond the reference point's scope. 
		\item [4)] 
		DMO demonstrates remarkable stability compared to other algorithms, which exhibit considerable variation from run to run due to their inability to consistently handle integer constraints. 
	\end{itemize}
	
	\textcolor{black}{Among the compared algorithms, GMOEA exhibits the best performance but also has the longest runtime. The performance of WOF-NSGA-II is relatively unaffected by problem size. However, MOEA/D-LWS and MOEA/HD perform worse than NSGA-II, suggesting that these two decomposition-based MOEAs are not well-suited for addressing the gasoline blending scheduling problem. }
	
	\begin{figure}[htbp]
		\centering
		\subfigure[]{
			\begin{minipage}[b]{0.45\textwidth}
				\includegraphics[width=1\textwidth]{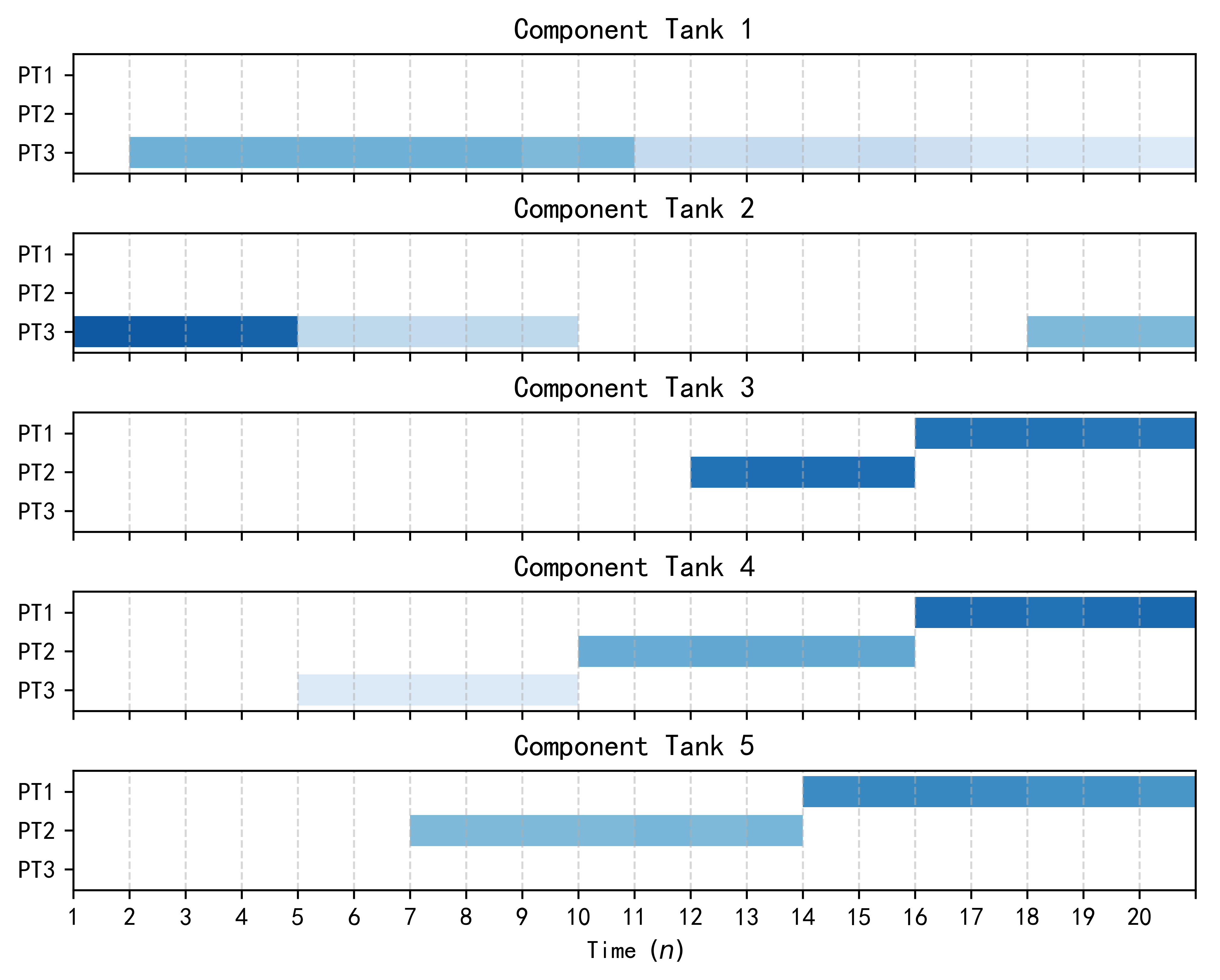}
			\end{minipage}\label{fig_schedule_DMO200}
		}
		\qquad
		\subfigure[]{
			\begin{minipage}[b]{0.45\textwidth}
				\includegraphics[width=1\textwidth]{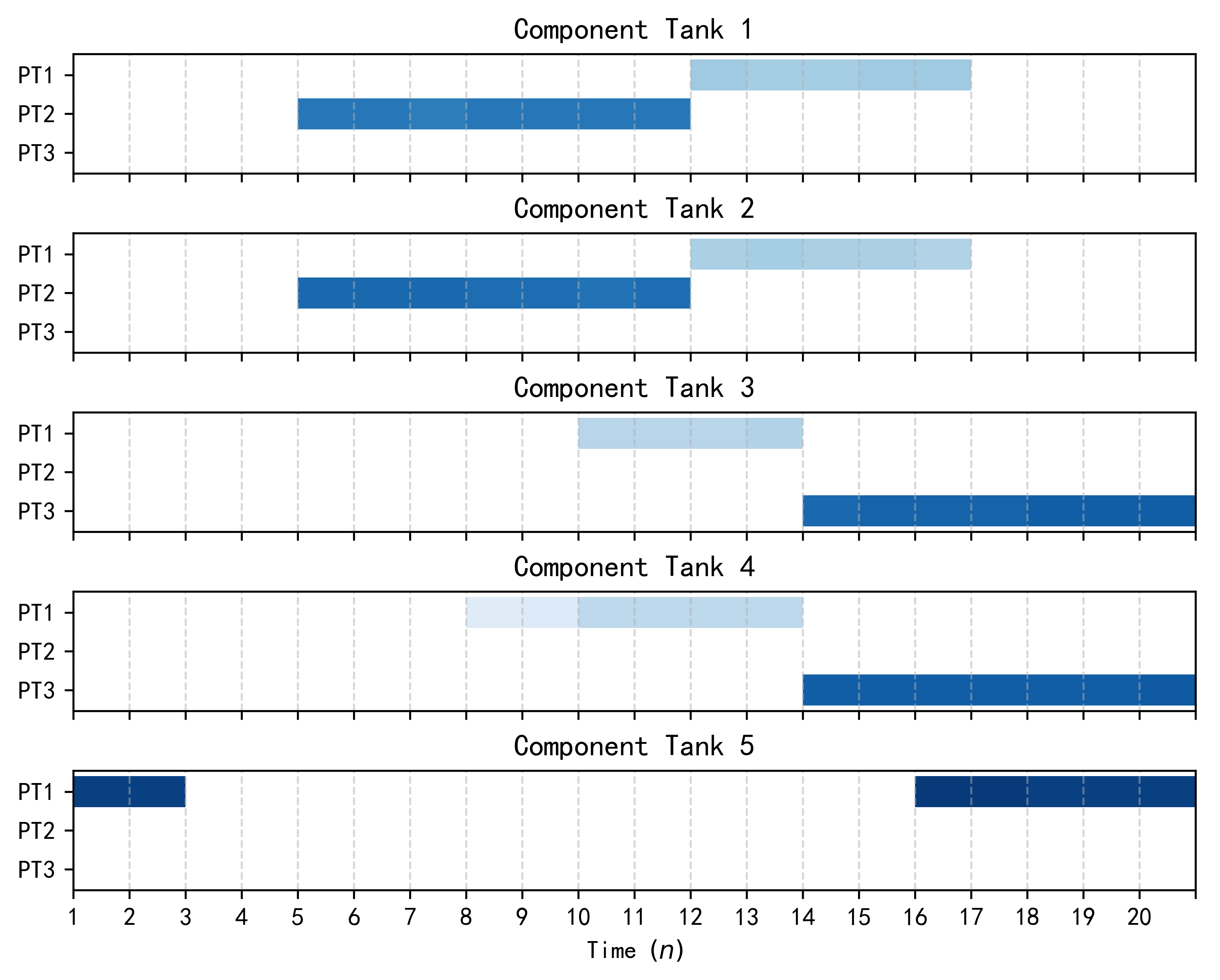}
			\end{minipage}\label{fig_schedule_DMO500}
		}
		\qquad
		\subfigure[]{
			\begin{minipage}[b]{0.45\textwidth}
				\includegraphics[width=1\textwidth]{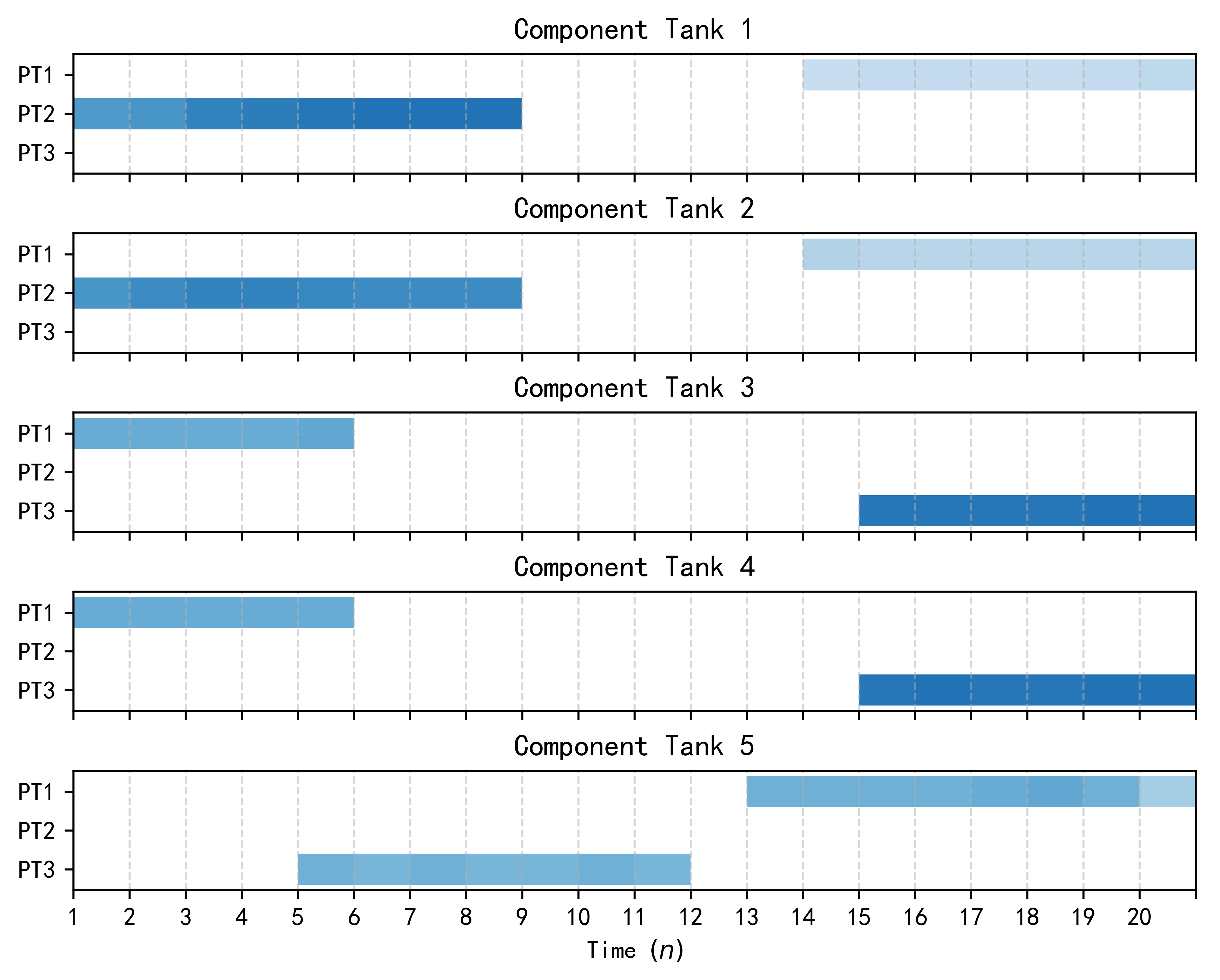}
			\end{minipage}\label{fig_schedule_DMO1000}
		}
		\caption{Examples of the Gantt charts of the schedules obtained by DMO for the gasoline blending scheduling problem with $N_{ct} = 5, N_{pt} = 3 \text{, and } n = 20$. The colored areas indicate that an oil component tank is transferring oil to product tank $j$ (vertical coordinate) during this specific period (horizontal coordinate). The shades of blue indicate the flow of oil transportation. (a) DMO $(T=200)$; (b) DMO $(T=500)$; (c) DMO $(T=1000)$. } 
		\label{fig_schedule_DMO}
	\end{figure} 
	
	Fig. \ref{fig_schedule_DMO} presents examples of Gantt charts illustrating the schedules obtained by DMO for the gasoline blending scheduling problem with $N_{ct} = 5, N_{pt} = 3 \text{, and } n = 20$. The schedule is based on a solution randomly selected from the Pareto front generated by DMO. It can be observed that DMO successfully generates executable Gantt charts for these schedules. Additionally, DMO tends to produce schedules with lower flow rates, which aligns with operators' preference \textcolor{black}{for using} moderate flow rates to extend the lifespan of pumps. This preference is learned by DMO from historical data. These results demonstrate \textcolor{black}{the ability of DMO} to emulate operator preferences, which are often difficult to translate into objectives and constraints. Consequently, the generated schedules exhibit greater consistency with the actual needs of \textcolor{black}{a} refinery. 
	
	According to the performance of the other compared algorithms \textcolor{black}{for} refineries, the proposed DMO can reduce ON waste by more than 0.5. Based on the refinery's production and the market price of gasoline, implementing DMO could lead to cost savings of at least \$20 million per year for the refinery used as the data source in this paper. 
	
\begin{figure}[tp]
		\centering
		\includegraphics[width=0.49\textwidth]{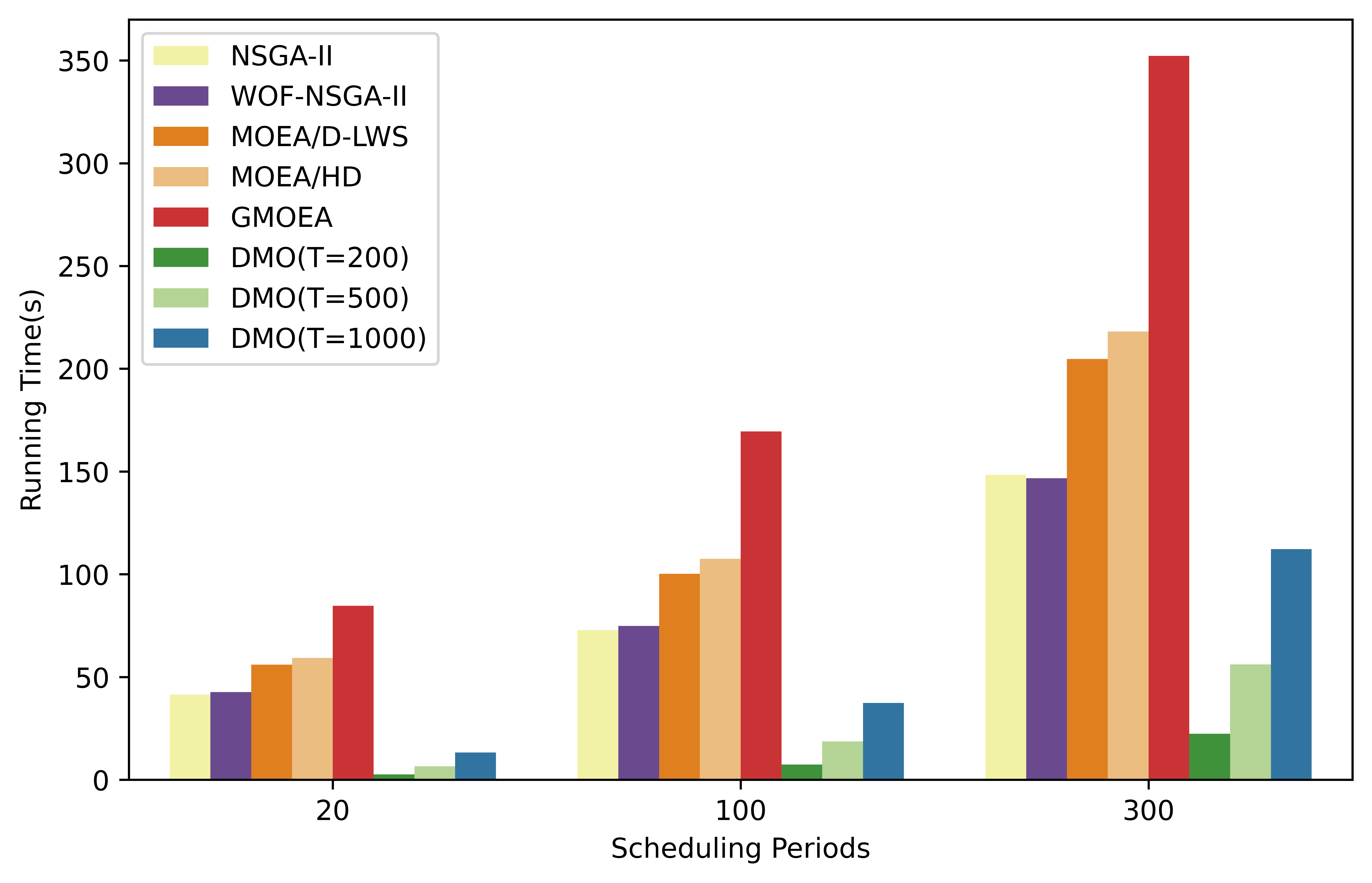}\\
		\color{black}\caption{The runtime results achieved by NSGA-II, WOF-NSGA-II, MOEA/D-LWS, MOEA/HD, GMOEA, and DMO ($T=200/500/1000$). }
		\label{fig_rt}
	\end{figure}

	\textcolor{black}{The running times of DMO ($T=200/500/1000$) and the other algorithms on the gasoline blending scheduling problem are also compared, with $N_{ct} = 5$, $N_{pt} = 3$, and $n = 20/100/300$, as shown in Fig. \ref {fig_rt}.} The results indicate that DMO has significantly shorter running times \textcolor{black}{than} the other algorithms. This characteristic aligns well with the requirements of refineries, which often \textcolor{black}{require} repeated adjustments and multiple runs. DMO achieves this efficiency by leveraging the powerful parallel computing capabilities of GPUs. Among the compared algorithms, GMOEA, which also employs GPU computing, exhibits the longest running time due to extensive data interaction between the GPU and CPU. As the problem size increases, the running time of DMO also increases proportionally. This suggests that the computational load reaches the upper limit of parallel computing on the Nvidia GeForce RTX 3090 GPU. Notably, the time complexity of DMO scales proportionally to the size of the processed schedule images because it utilizes convolutional neural networks. 
	
	\subsection{\textcolor{black}{Optimization Effectiveness of the Proposed DMO}} 
	\textcolor{black}{Experiments were also conducted to ascertain whether the optimization mechanisms of DMO contribute to its exceptional performance, rather than other factors.} Ablation experiments were carried out by comparing DMO with random generators (diffusion models without optimization) to validate the effectiveness of DMO. In this experiment, the T value for DMO was set to 500 for representativeness. 
	
	\begin{table}[htbp] 
		\color{black}\caption{HV Results Obtained by DMO ($T=500$) and Random Generator on the Gasoline Blending Scheduling Problem with (a) $N_{ct} = 5, N_{pt} = 3 \text{, and } n = 20$; (b) $N_{ct} = 8, N_{pt} = 5 \text{, and } n = 100$; (c) $N_{ct} = 12, N_{pt} = 7 \text{, and } n = 300$. The Best Result is Bolded. The symbols ``$+$", ``$-$", and ``$\approx$" Indicate that the Algorithm is Significantly better than,  worse than, and Statistically Tied with the Other Method. }
		\label{tab_abla}
		\centering
		\begin{tabular}{ccc}
			\toprule
			& DMO                      & Random                   \\
			\midrule
			(a)                & \textbf{0.979(1.124e-5)} & 0.215(8.937e-4)          \\
			(b) 			   & \textbf{0.842(2.993e-5)} & 0.062(1.301e-4)          \\
			(c) 			   & \textbf{0.726(5.889e-5)} & 0.083(1.974e-4)          \\
			\midrule
			$+/-/\approx$	   &                          & $0/3/0$                  \\
			\bottomrule         
		\end{tabular}\\
	\end{table}
	
	\begin{figure}[htbp]
		\centering
		\subfigure[]{
			\begin{minipage}[b]{0.45\textwidth}
				\includegraphics[width=1\textwidth]{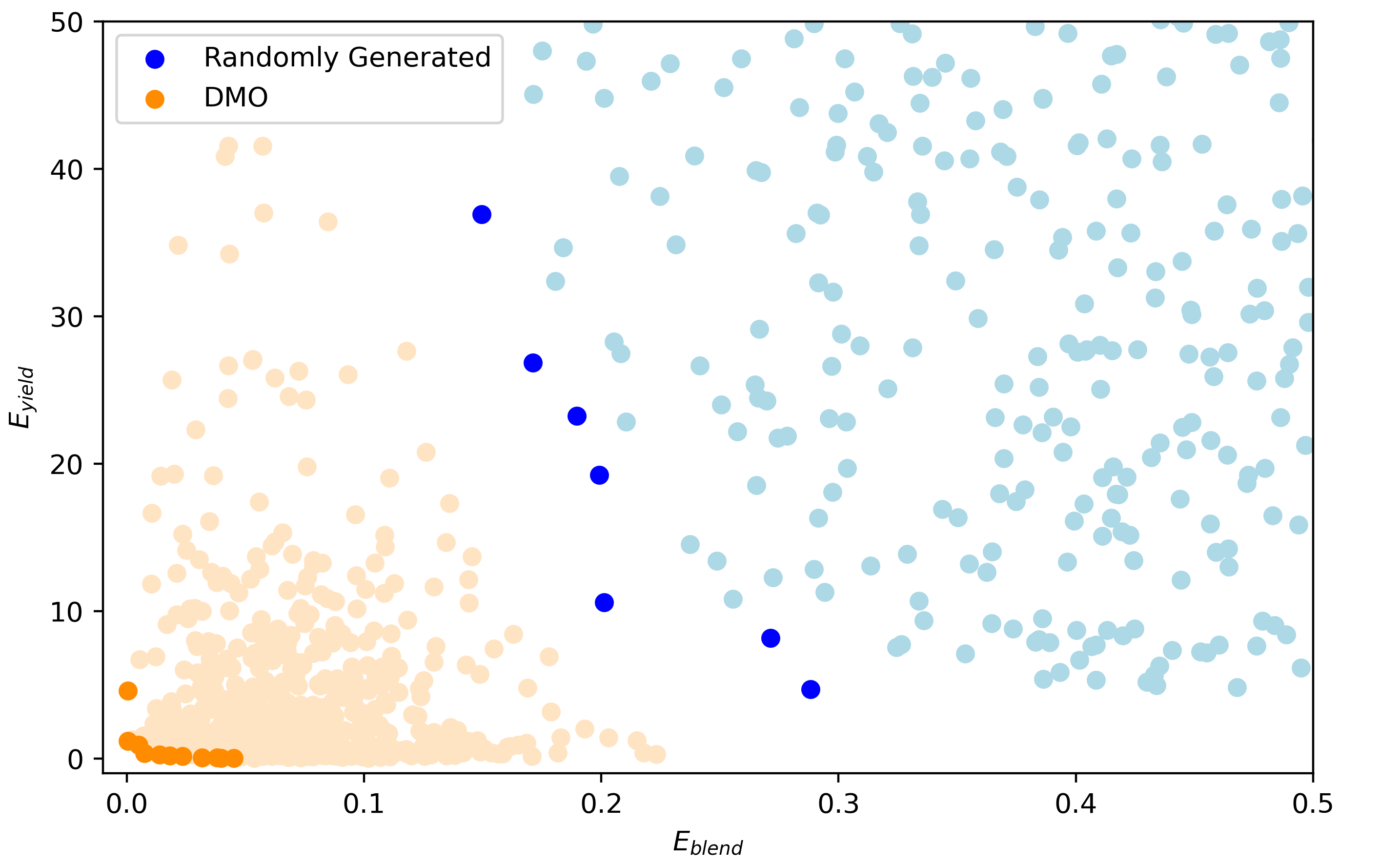}
			\end{minipage}\label{fig_compare_ifopt1}
		}
		\qquad
		\subfigure[]{
			\begin{minipage}[b]{0.45\textwidth}
				\includegraphics[width=1\textwidth]{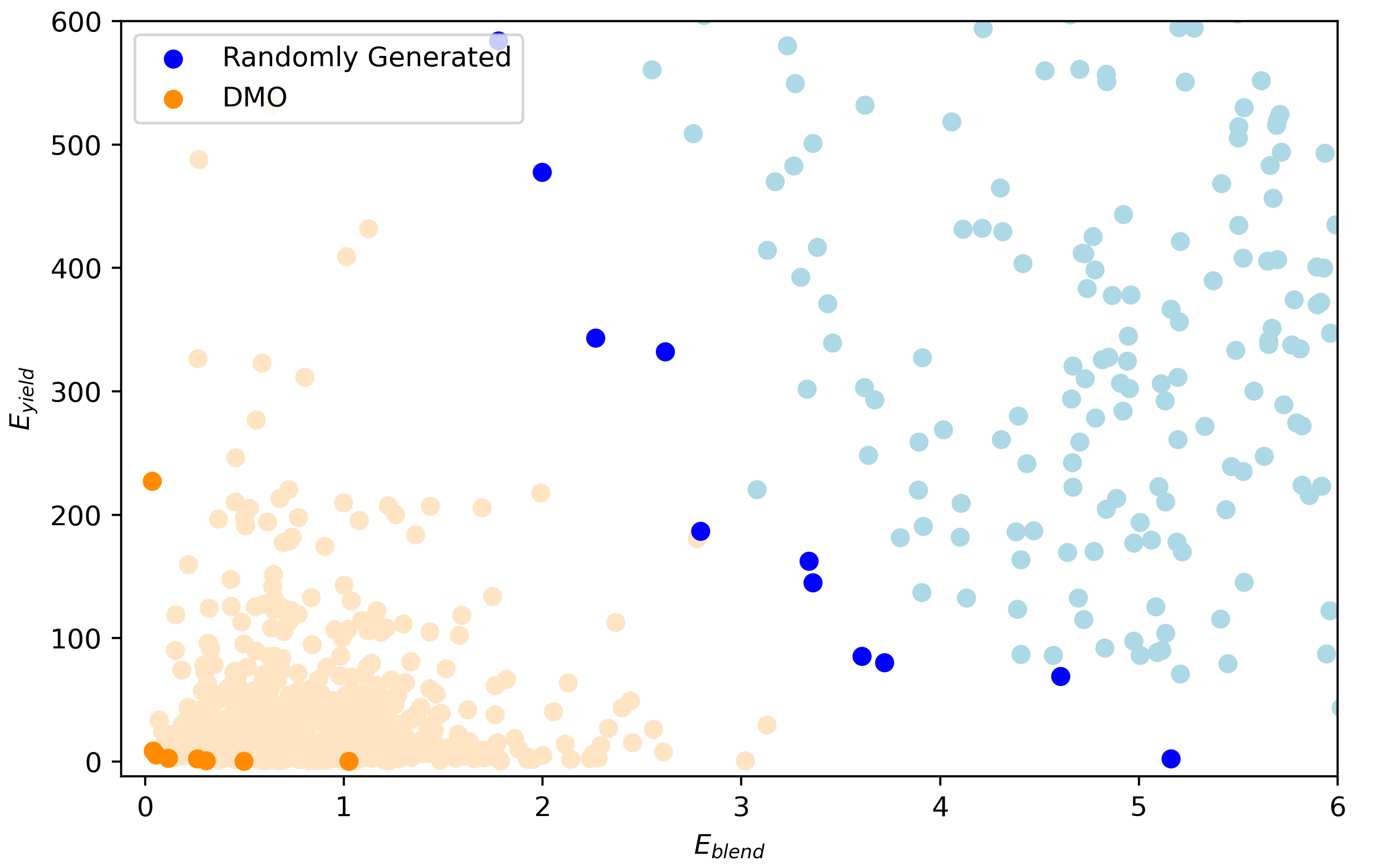}
			\end{minipage}\label{fig_compare_ifopt2}
		}
		\qquad
		\subfigure[]{
			\begin{minipage}[b]{0.45\textwidth}
				\includegraphics[width=1\textwidth]{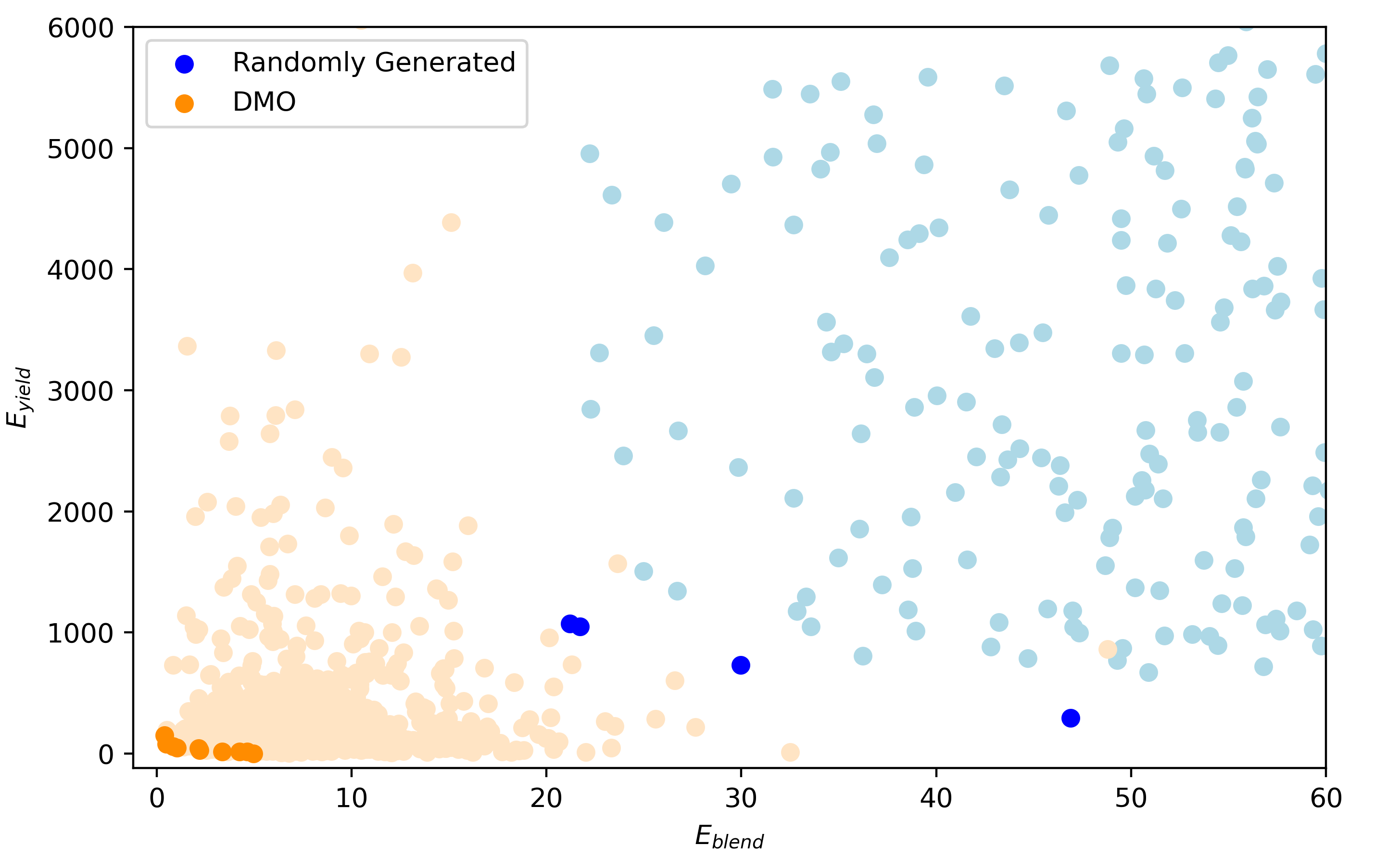}
			\end{minipage}\label{fig_compare_ifopt3}
		}
		\color{black}\caption{The final population obtained by DMO ($T=500$) and a random generator on the gasoline blending scheduling problem in the objective space. The dark-colored points are \textcolor{black}{nondominated} solutions. (a) $N_{ct} = 5, N_{pt} = 3 \text{, and } n = 20$; (b) $N_{ct} = 8, N_{pt} = 5 \text{, and } n = 100$; (c) $N_{ct} = 12, N_{pt} = 7 \text{, and } n = 300$. } 
		\label{fig_compare_ifopt}
	\end{figure} 
	
	The results displayed in Table \ref{tab_abla} reveal that DMO significantly outperforms the random generator, thereby confirming the effectiveness of \textcolor{black}{the DMO optimization mechanism}. Furthermore, the visual comparison in Fig. \ref{fig_compare_ifopt} illustrates the significant difference in performance between the two solutions. 
	It is evident that DMO excels in producing a considerable number of feasible solutions; however, it is noteworthy that not all solutions generated by DMO are necessarily Pareto optimal. This is because each solution in DMO is generated independently of the others. Notably, DMO lacks a selection mechanism ensuring that all final population solutions are \textcolor{black}{nondominated}. We \textcolor{black}{aimed} to design a selection mechanism for DMO akin to \textcolor{black}{that of} MOEA/D \cite{MOEAD}. However, due to the incomparability of objectives during the DMO run (\textcolor{black}{for which solutions are not feasible}), the selection mechanism did not yield the desired effect. \textcolor{black}{Instead, it rendered a decline of quality of the final solutions. We aim to pursue this line of research in our future endeavors. \textcolor{black}{Even though} only a few solutions of DMO are \textcolor{black}{nondominated}, it continues to perform exceptionally well. Moreover, as DMO generates each solution independently, the central limit theorem ensures the stability of DMO. }
	
	\subsection{Optimization Behavior of the Proposed DMO} 
	Since diffusion models have not been previously used in practical optimization problems, it is necessary to investigate how DMO performs \textcolor{black}{in} optimization. This investigation involves observing the solutions generated during algorithm runs. To easily observe the variation in a specific solution, a problem size of $N_{ct} = 5, N_{pt} = 3 \text{, and } n = 20$ was chosen. 
	
	\begin{figure}[htbp]
		\centering
		\includegraphics[width=0.49\textwidth]{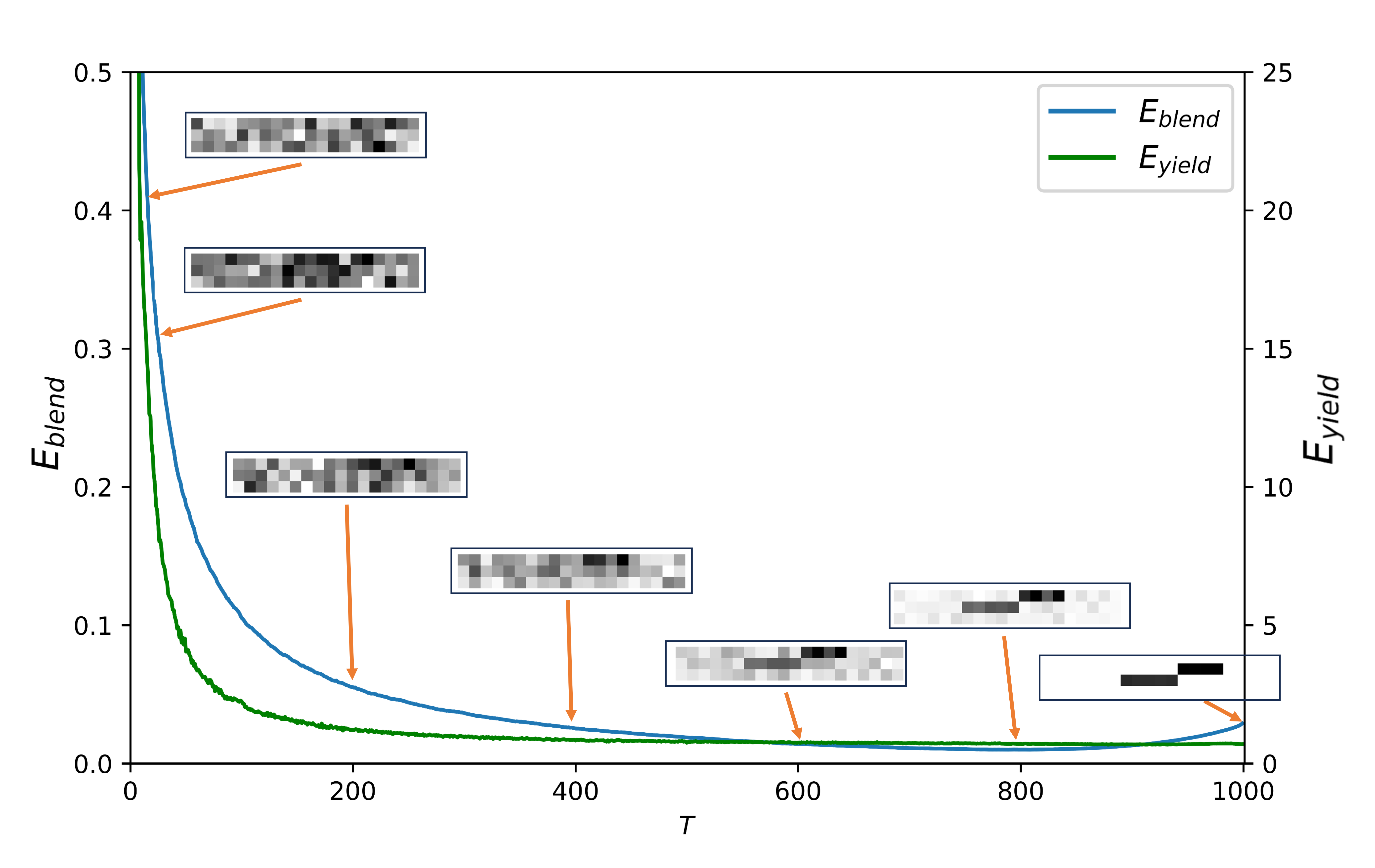}\\
		\caption{An example of the optimization behavior of DMO on the gasoline blending scheduling problem, in which $N_{ct} = 5, N_{pt} = 3 \text{, and } n = 20$. 
		}
		\label{fig_DMOrunning}
	\end{figure}
	
	Fig. \ref{fig_DMOrunning} illustrates the variation in the mean value of the two objectives as the number of iterations ($T$) increases. It is evident that both objectives generally decrease with \textcolor{black}{an increasing number of DMO iterations}, indicating the feasibility of using gradient descent to directly optimize the objectives. Despite the contradictory nature of the two objectives in the feasible domain, DMO can simultaneously optimize both objectives by generating multiple intermediate distributions between the Gaussian noise and the feasible domain.
	Furthermore, towards the end of the algorithm \textcolor{black}{process}, both objectives \textcolor{black}{increases slightly}, which can be attributed to the final correction of solutions by the diffusion model. The diffusion model ensures that the partial solutions conform to the constraints as well as possible, resulting in a slight increase on the objectives. 
	Fig. \ref{fig_DMOrunning} also displays a specific solution depicted as part of the solution, which gradually transitions from Gaussian noise to a clear schedule as the algorithm iterates. This observation supports the conclusion that DMO is capable of optimizing the objectives while gradually aligning the solutions with the given constraints. 
	
	\subsection{\textcolor{black}{Advantages and Limitations of the Proposed DMO}} 
	\textcolor{black}{This section outlines the advantages and limitations of DMO}. The advantages are summarized as follows:
	\begin{itemize}
		\item [1)]
		DMO facilitates the transition of solutions from Gaussian noise to feasible solutions using a diffusion model while optimizing \textcolor{black}{the} objectives through gradient descent. This unique optimization mechanism empowers DMO to significantly outperform \textcolor{black}{the} other compared algorithms \textcolor{black}{when} solving the gasoline blending scheduling problem. Moreover, its efficiency is notably enhanced by its effective utilization of GPU computing capabilities.
		\item [2)]
		DMO exhibits excellent \textcolor{black}{generalizability} and scalability. It can handle gasoline blending scheduling problems of varying sizes without requiring retraining. When applied to other scheduling optimization problems, modifications are only necessary for the training data and for the objectives optimized via gradient descent. Furthermore, by transforming sequences into adjacency graphs, DMO holds promise for addressing combinatorial optimization problems.
	\end{itemize}
	
	\textcolor{black}{However, DMO also has certain limitations: } 
	\begin{itemize}
		\item [1)]
		As each solution generated by DMO is independent of \textcolor{black}{the} others, a large number of \textcolor{black}{the} solutions produced may not be Pareto optimal. This inefficiency in solution generation might lead to significant computational waste. Future enhancements could involve devising a selection mechanism to optimize algorithmic performance for the population. 
		\item [2)]
		DMO generates new scheduling solutions by learning from historical data, which tends to align \textcolor{black}{the} solutions more closely with operational habits. Nevertheless, this adherence to specific data distributions can limit the optimization potential \textcolor{black}{of DMO}. This limitation \textcolor{black}{could} be mitigated by adjusting the training data in future iterations. 
	\end{itemize}
	
	\section{Conclusion}
	In this paper, a diffusion model-driven \textcolor{black}{multiobjective} optimization method (DMO) is proposed, \textcolor{black}{which is} specifically designed to \textcolor{black}{address} the gasoline blending scheduling problem. \textcolor{black}{By leveraging the generative abilities of the diffusion model}, DMO proves to be highly efficient in solving this practical optimization problem. An innovative approach is introduced that utilizes the diffusion model to generate feasible solutions that adhere to integer constraints. During diffusion model iteration, the objectives are simultaneously optimized using gradient descent. By creating multiple intermediate distributions between Gaussian noise and the feasible domain, DMO optimizes \textcolor{black}{the} objectives while ensuring constraint compliance. This novel optimization strategy enables DMO to explore the solution space more efficiently. Additionally, due to the use of convolutional neural networks as models, DMO exhibits excellent performance in handling large-scale problems.
	
	To assess the performance of DMO, empirical comparisons were conducted on gasoline blending scheduling problems with up to 50,400 decision variables. The results consistently \textcolor{black}{demonstrated} the superiority of the DMO algorithm over three compared MOEAs: NSGA-II, WOF-NSGA-II, MOEA/D-LWS, MOEA/HD, and GMOEA. DMO achieves the best and most stable results while maintaining its status as the most efficient algorithm. 
	
	This paper demonstrates the promising potential of DMO in solving the gasoline blending scheduling problem. The \textcolor{black}{adaptable} framework of DMO enables straightforward extension to other scheduling problems by simply adjusting the training data and objective function. This flexibility warrants further exploration to expand the application \textcolor{black}{of DMO} to optimization problems characterized by large decision spaces, intricate constraints, and dynamic environments. Furthermore, DMO demonstrates significant promise in addressing combinatorial optimization problems. 
	
	\section{Acknowledgment}
	This work was supported by National Natural Science Foundation of China (Key Program: 62136003), National Natural Science Foundation of China (62173144,62073142), National Natural Science Foundation of Shanghai (21ZR1416100), the Programme of Introducing Talents of Discipline to Universities (the 111 Project) under Grant B17017 and Shanghai AI lab. 
	
	\ifCLASSOPTIONcaptionsoff
	\newpage
	\fi
	
	{\footnotesize\bibliography{reference/ref}
		\bibliographystyle{ieeetr}}

\end{document}